\documentclass[journal]{IEEEtran}
\ifCLASSINFOpdf
\else
\fi

\usepackage[table,xcdraw]{xcolor}
\usepackage{amsmath}
\usepackage{epsfig}
\usepackage{graphics}
\usepackage{multirow}
\usepackage{subcaption}
\usepackage{mwe}
\usepackage{makecell}
\usepackage{url}

\newcommand{\comment}[1]{}

\usepackage[normalem]{ulem} 

\usepackage{xspace}

\usepackage{hyperref} 
\usepackage{bm} 
\usepackage{amsmath} 
\usepackage{amssymb}
\usepackage{cleveref}

\usepackage{adjustbox} 
\usepackage{multirow}
\usepackage{booktabs}
\usepackage[]{pgfplots} 
\usepackage[]{tikz}
\pgfplotsset{compat=1.9}

\usepackage{algorithmicx}
\usepackage{algorithm}
\usepackage[noend]{algpseudocode}

\usepackage{color}

\let\oldtextbf\textbf
\renewcommand{\textbf}[1]{\oldtextbf{\boldmath #1}}

\begin{document}
%
\title{SEN12MS-CR-TS: A Remote Sensing Data Set for Multi-modal Multi-temporal Cloud Removal}
%
%
%

\author{Patrick~Ebel, 
        Yajin~Xu,~
        Michael~Schmitt,~\IEEEmembership{Senior Member,~IEEE,}
        and~Xiao~Xiang~Zhu,~\IEEEmembership{Fellow,~IEEE}

\thanks{This work was partially supported by the Federal Ministry for Economic Affairs and Energy of Germany in the project “AI4Sentinels– Deep Learning for the Enrichment of Sentinel Satellite Imagery” (FKZ50EE1910). The work of X. Zhu is jointly supported by the European Research Council (ERC) under the European Union's Horizon 2020 research and innovation programme (grant agreement No. [ERC-2016-StG-714087], Acronym: \textit{So2Sat}), by the Helmholtz Association
through the Framework of Helmholtz AI (grant  number:  ZT-I-PF-5-01) - Local Unit ``Munich Unit @Aeronautics, Space and Transport (MASTr)'' and Helmholtz Excellent Professorship ``Data Science in Earth Observation - Big Data Fusion for Urban Research''(grant number: W2-W3-100), by the German Federal Ministry of Education and Research (BMBF) in the framework of the international future AI lab "AI4EO -- Artificial Intelligence for Earth Observation: Reasoning, Uncertainties, Ethics and Beyond" (grant number: 01DD20001) and by German Federal Ministry of Economics and Technology in the framework of the "national center of excellence ML4Earth" (grant number: 50EE2201C).\\
(\textit{Corresponding Author: Xiao Xiang Zhu})} 

\thanks{P. Ebel and Y. Xu are with Data Science in Earth Observation, Technical
University of Munich, 80333 Munich, Germany. (e-mail: patrick.ebel@tum.de).}
\thanks{M. Schmitt is with the Remote Sensing Technology Institute, German Aerospace Center, 82234 Wessling, Germany, and also with the Chair of Earth Observation, Bundeswehr University Munich, 85577 Neubiberg, Germany . (e-mail: michael.schmitt@unibw.de).}
\thanks{X. X. Zhu is with Data Science in Earth Observation, Technical University of Munich, 80333 Munich, Germany and also with the Remote Sensing Technology Institute, German Aerospace Center, 82234 Wessling, Germany. (e-mail: xiaoxiang.zhu@dlr.de).}

}
%
%

\markboth{IEEE TRANSACTIONS ON GEOSCIENCE AND REMOTE SENSING, , In press.}
{Shell \MakeLowercase{\textit{et al.}}: Bare Demo of IEEEtran.cls for Journals}
%



\maketitle

\begin{abstract}

    About half of all optical observations collected via spaceborne satellites are affected by haze or clouds. Consequently, cloud coverage affects the remote sensing practitioner's capabilities of a continuous and seamless monitoring of our planet. This work addresses the challenge of optical satellite  image reconstruction and cloud removal by proposing a novel multi-modal and multi-temporal data set called SEN12MS-CR-TS. We propose two models highlighting the benefits and use cases of SEN12MS-CR-TS: First, a multi-modal multi-temporal 3D-Convolution Neural Network that predicts a cloud-free image from a sequence of cloudy optical and radar images. Second, a sequence-to-sequence translation model that predicts a cloud-free time series from a cloud-covered time series. Both approaches are evaluated experimentally, with their respective models trained and tested on SEN12MS-CR-TS. The conducted experiments highlight the contribution of our data set to the remote sensing community as well as the benefits of multi-modal and multi-temporal information to reconstruct noisy information. 
    Our data set is available at \textit{\url{https://patrickTUM.github.io/cloud_removal}}. 
\end{abstract}


%
\IEEEpeerreviewmaketitle


\section{Introduction}

\IEEEPARstart{T}{he} majority of our planet's land surface is covered by haze or clouds \cite{King_Platnick_Menzel_Ackerman_Hubanks_2013}. Such atmospheric distortions impede the capability of spaceborne optical satellites to reliably and seamlessly record noise-free data of the Earth's surface. The presence of clouds is detrimental to typical remote sensing applications, for instance land cover classification \cite{schmitt2021remote}, semantic segmentation \cite{Rafique_Blanton_Jacobs, schmitt2020weakly} and change detection \cite{ebel2021fusing, saha2021self}.

\begin{figure}[h!tb]
  \centering
  \begin{subfigure}[b]{0.48\linewidth}
    \includegraphics[width=\linewidth]{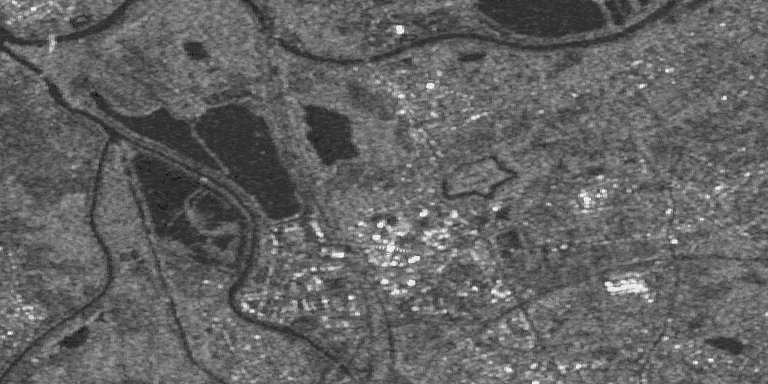}
  \end{subfigure}
    \begin{subfigure}[b]{0.48\linewidth}
    \includegraphics[width=\linewidth]{Fig/ROI_teaser/001_sar.png}
  \end{subfigure}
  
  \begin{subfigure}[b]{0.48\linewidth}
    \includegraphics[width=\linewidth]{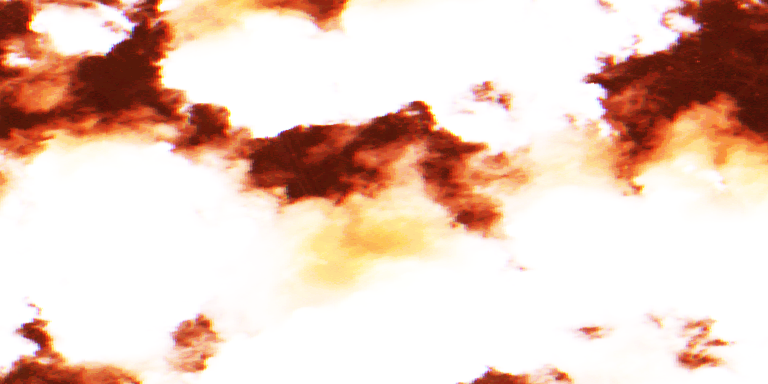}
  \end{subfigure}
    \begin{subfigure}[b]{0.48\linewidth}
    \includegraphics[width=\linewidth]{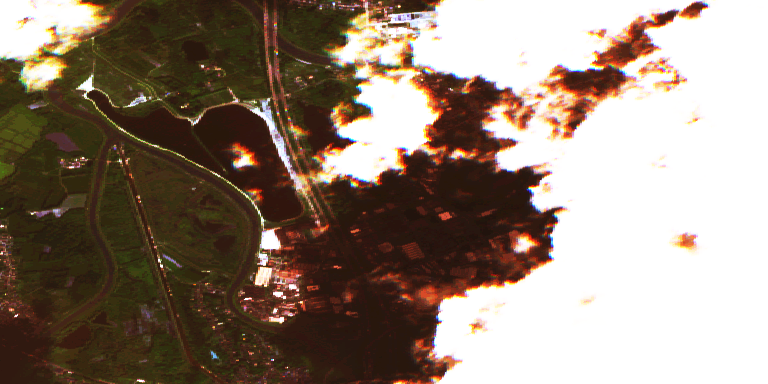}
  \end{subfigure}

  \begin{subfigure}[b]{0.48\linewidth}
    \includegraphics[width=\linewidth]{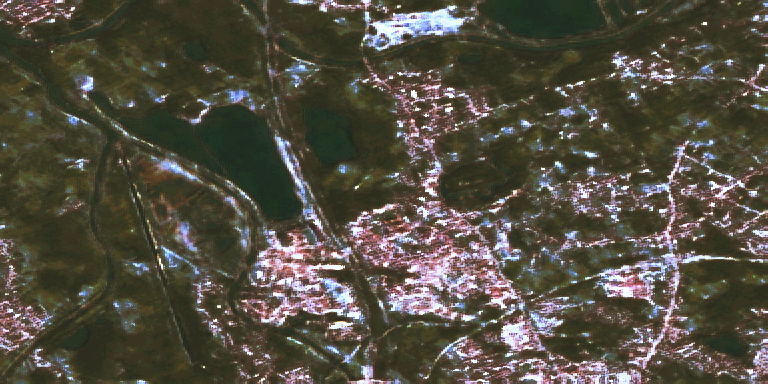}
  \end{subfigure}
    \begin{subfigure}[b]{0.48\linewidth}
    \includegraphics[width=\linewidth]{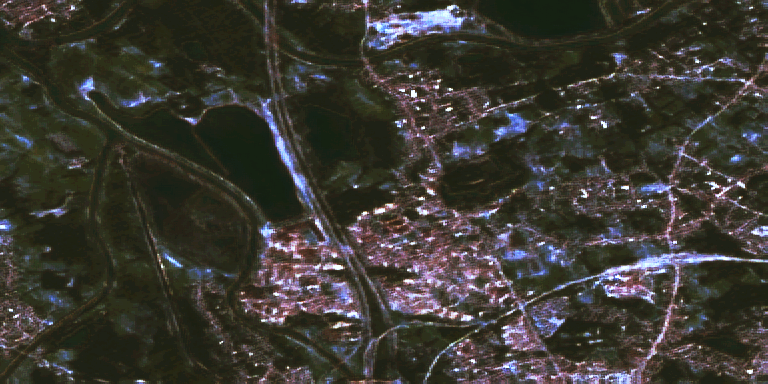}
  \end{subfigure}
  
  \begin{subfigure}[tb]{0.975\linewidth}
    \includegraphics[width=\linewidth]{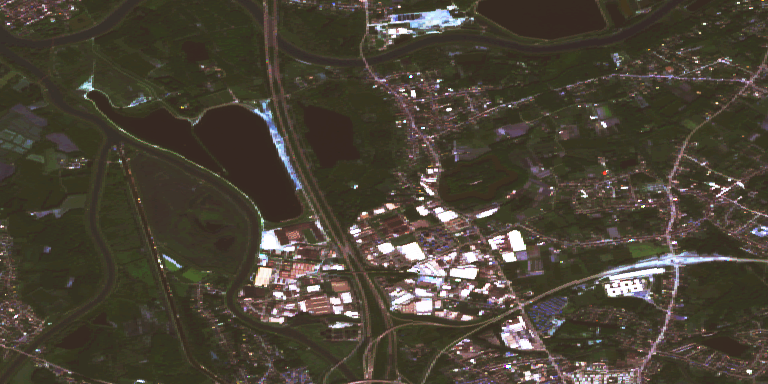}
  \end{subfigure}
  \caption{Example observations and cloud-free predictions. Columns: Samples at two different time points. Rows: S1 data (in grayscale), cloudy S2 data (in RGB), predicted cloud-free $\hat{S2}$ data, reference cloud-free S2 data of a later point in time. The results highlight that our network is able to integrate multi-modal and multi-temporal information to predict a clear-view sequence of multi-spectral observations, even in the presence of heavy cloud coverage.}
  \label{fig:teaser}
\end{figure}

Hence, the need for cloud-free earth observation gave rise to a rapidly growing number of haze and cloud removal methods \cite{Enomoto_Sakurada_Wang_Fukui_Matsuoka_Nakamura_Kawaguchi_2017, Grohnfeldt_Schmitt_Zhu_2018, Singh_Komodakis_2018,  Bermudez_Happ_Feitosa_Oliveira_2019, Rafique_Blanton_Jacobs, gu2019single, Sarukkai_Jain_Uzkent_Ermon_2019, meraner2020cloud, ebel2020multisensor}. Most previous methods focus on a multi-modal approach \cite{Grohnfeldt_Schmitt_Zhu_2018, ebel2020cloud, meraner2020cloud, ebel2020multisensor} to reconstruct cloud-covered pixels via information translated from synthetic aperture radar (SAR) or other sensors more robust to atmospheric disturbances \cite{bamler2000principles}, yet focus on only a single time-point of observations. In comparison, recent models attempt a temporal reconstruction of cloudy observations by means of inference across time-series \cite{Sarukkai_Jain_Uzkent_Ermon_2019, oehmcke2020creating, zhang2021combined}, utilizing the circumstance that the extent of cloud coverage over a particular region is variable over time and seasons \cite{King_Platnick_Menzel_Ackerman_Hubanks_2013}.

\begin{figure*}[h!tb]
    \centering  
    \includegraphics[width=\linewidth]{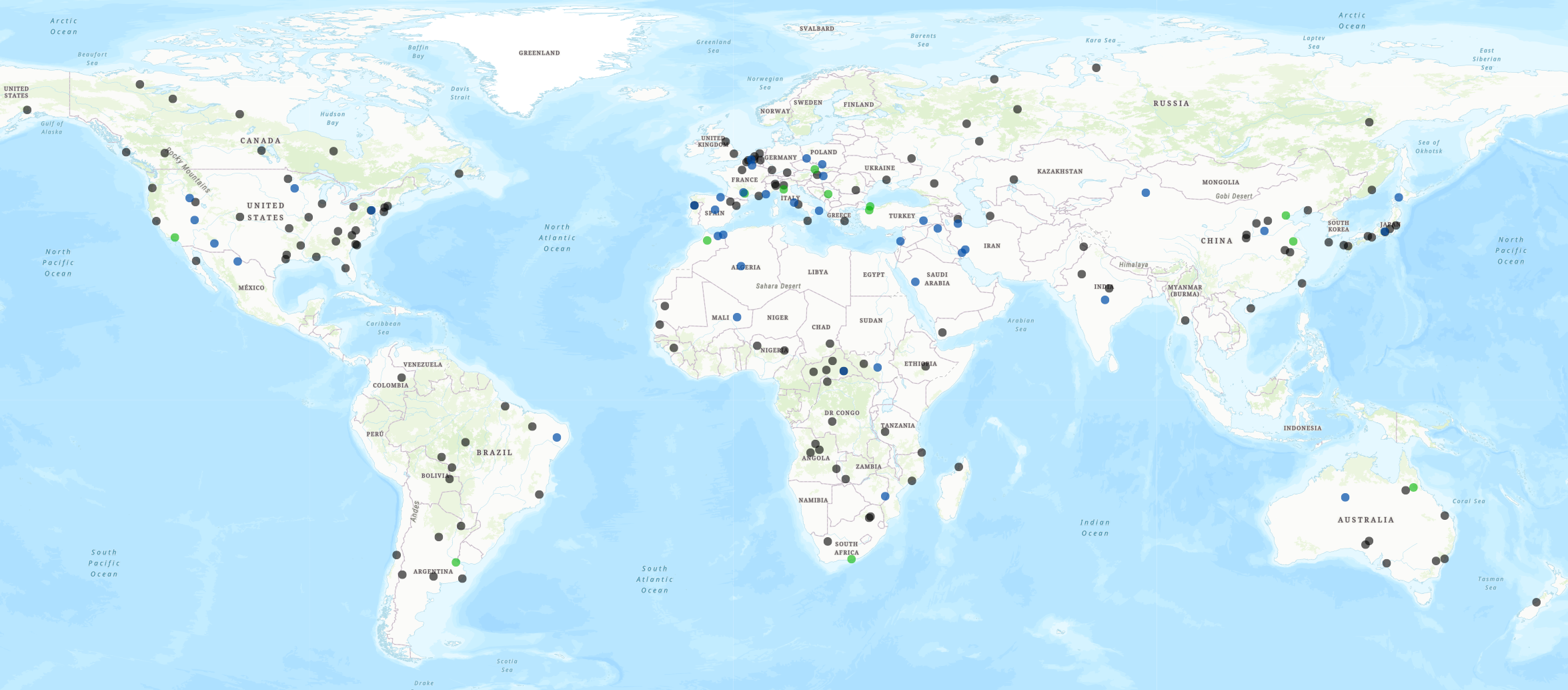}
    \caption{Spatial distribution of the ROI constituting SEN12MS-CR-TS. Areas belonging to the training split are denoted in blue, regions of the testing split are colored in green. The ROI of SEN12MSCR \cite{ebel2020multisensor}, non-overlapping and compatible with our data set, are depicted in gray. Any graphical overlap of the semi-transparently plotted dots is rendered in darker tones so close-by dots can easier be discerned.}
    \label{fig:map}
\end{figure*}

The work at hand aims to combine both preceding approaches and thus considers the challenge of cloud removal in optical satellite imagery by integrating information across time and within different modalities. For this purpose, we curate a new data set called SEN12MS-CR-TS, which contains multi-temporal and multi-modal satellite observations. Specifically, SEN12MS-CR-TS consists of 1-year long time-series of co-registered radar Sentinel-1 (S1) as well as multi-spectral Sentinel-2 observations (S2) acquired in a paired manner, covering regions of interest (ROI) from all over the world. We highlight the benefits of the proposed data set by training and testing two different models on our data set: First, a multi-modal multi-temporal 3D-Convolution Neural Network that predicts a cloud-free image from a sequence of cloudy optical and radar images. Second, a sequence-to-sequence translation model that predicts a cloud-free time series from a cloud-covered time series. Both approaches are evaluated experimentally, with their respective models trained and tested on SEN12MS-CR-TS. The conducted experiments highlight the contribution of our curated data set to the remote sensing community as well as the benefits of multi-modal and multi-temporal information to reconstruct noisy information.

\begin{figure*}[h!tb]
  \centering
  \begin{subfigure}[b]{0.19\linewidth}
    \includegraphics[width=\linewidth]{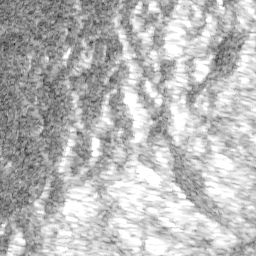}
  \end{subfigure}
  \begin{subfigure}[b]{0.19\linewidth}
    \includegraphics[width=\linewidth]{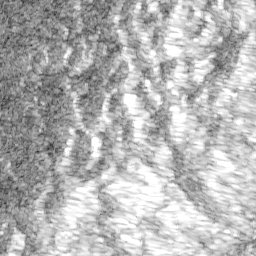}
  \end{subfigure}
    \begin{subfigure}[b]{0.19\linewidth}
    \includegraphics[width=\linewidth]{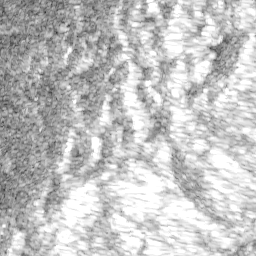}
  \end{subfigure}
      \begin{subfigure}[b]{0.19\linewidth}
    \includegraphics[width=\linewidth]{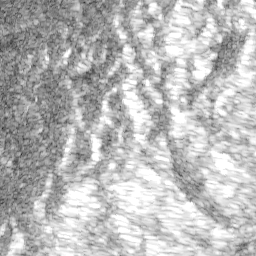}
  \end{subfigure}
      \begin{subfigure}[b]{0.19\linewidth}
    \includegraphics[width=\linewidth]{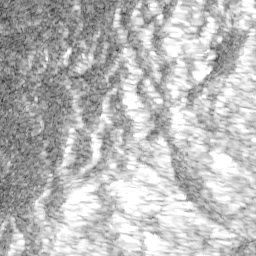}
  \end{subfigure}
  \begin{subfigure}[b]{0.19\linewidth}
    \includegraphics[width=\linewidth]{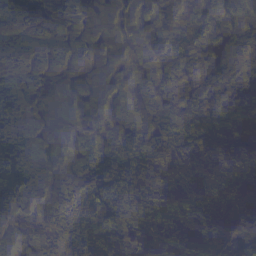}
  \end{subfigure}
  \begin{subfigure}[b]{0.19\linewidth}
    \includegraphics[width=\linewidth]{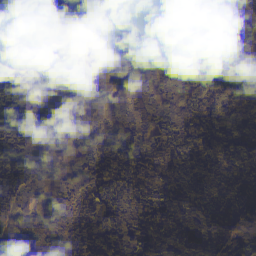}
  \end{subfigure}
    \begin{subfigure}[b]{0.19\linewidth}
    \includegraphics[width=\linewidth]{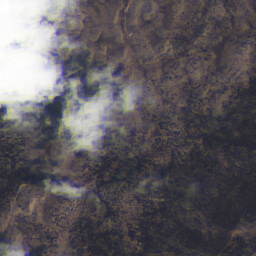}
  \end{subfigure}
      \begin{subfigure}[b]{0.19\linewidth}
    \includegraphics[width=\linewidth]{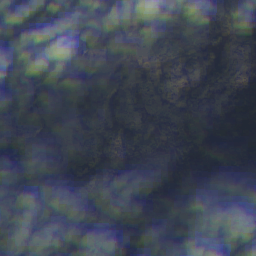}
  \end{subfigure}
      \begin{subfigure}[b]{0.19\linewidth}
    \includegraphics[width=\linewidth]{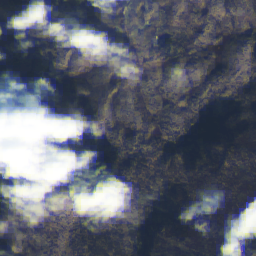}
  \end{subfigure}
  \begin{subfigure}[b]{0.19\linewidth}
    \includegraphics[width=\linewidth]{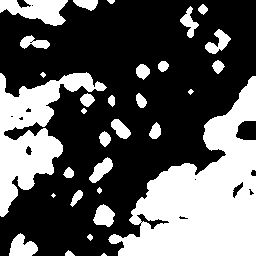}
  \end{subfigure}
  \begin{subfigure}[b]{0.19\linewidth}
    \includegraphics[width=\linewidth]{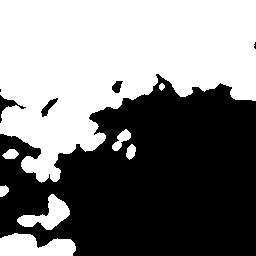}
  \end{subfigure}
    \begin{subfigure}[b]{0.19\linewidth}
    \includegraphics[width=\linewidth]{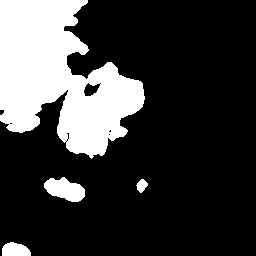}
  \end{subfigure}
      \begin{subfigure}[b]{0.19\linewidth}
    \includegraphics[width=\linewidth]{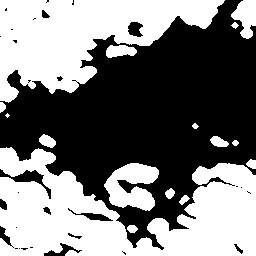}
  \end{subfigure}
      \begin{subfigure}[b]{0.19\linewidth}
    \includegraphics[width=\linewidth]{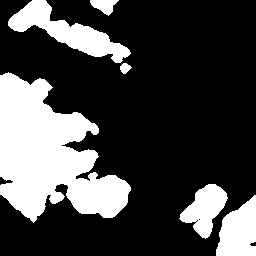}
  \end{subfigure}
  \caption{Example data, preprocessed as stated in section \ref{preprocessing}. Rows: S1 data (in grayscale), S2 data (in RGB), binary cloud masks (as per s2cloudless \cite{Zupanc}). Columns: Samples of five different time points. The illustrations show that the observed region is affected by variable atmospheric disturbances and covered by a dynamic extent of clouds, changing over time. The detected cloud coverage at the individual time points is 36, 49, 23, 48 percent, with an average of about 39 percent across all illustrated samples. While some pixels are clear at least at one point in the series and may thus be reconstructed by integrating across time, others are cloud-covered throughout the sequence and require spatial context or cloud-robust sensor information to be reconstructed.
  }
  \label{fig:sample_patches_masks}
\end{figure*}

\begin{figure}[h!tb]
    \includegraphics[width=\linewidth]{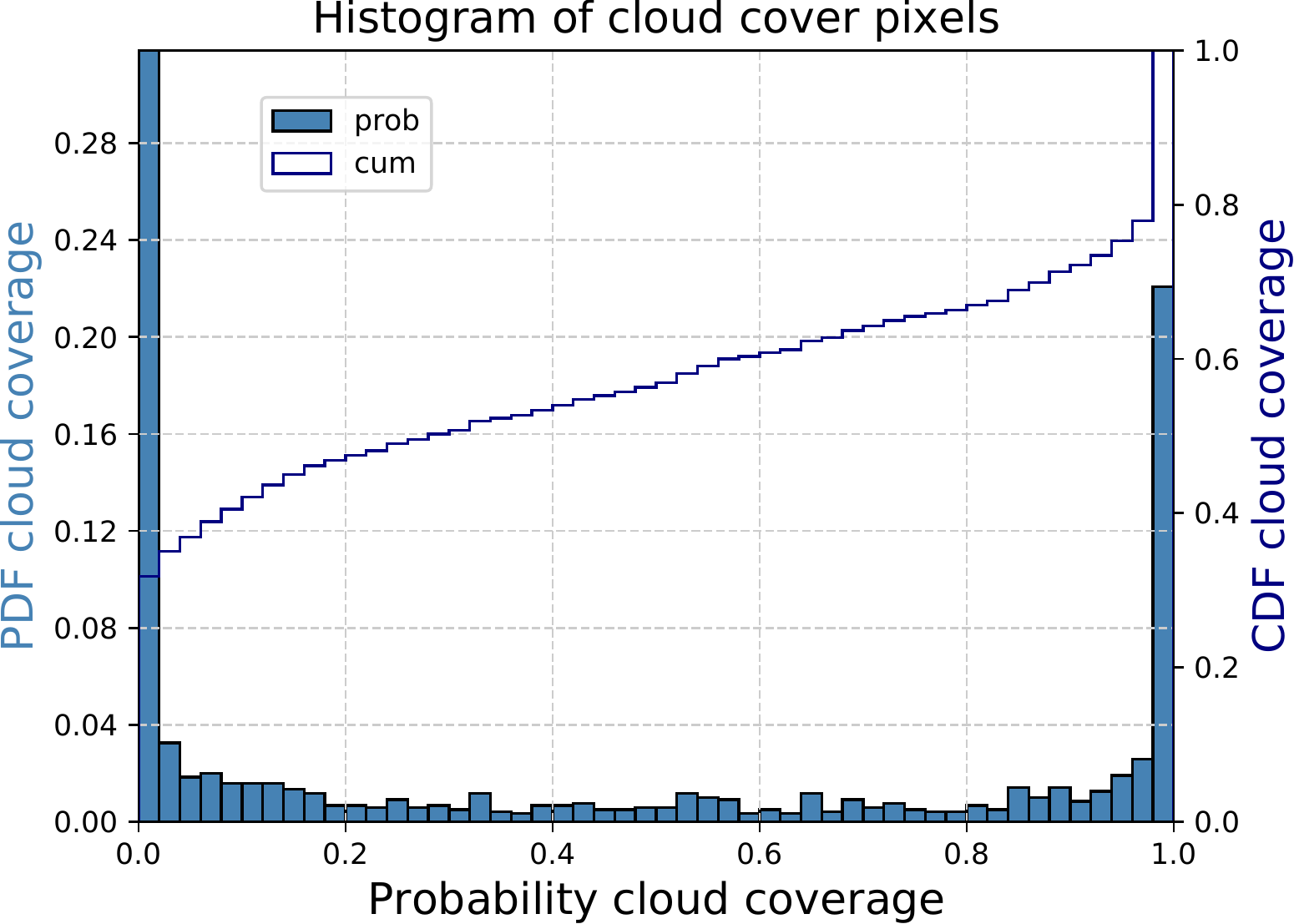}
    \caption{Statistics of cloud coverage of SEN12MS-CR-TS train split, computed on full-scene images via the detector of \cite{Zupanc}. On average, approximately $44 \%$ $(\pm 42 \%)$ of occlusion is observed. The empirical distribution of cloud coverage is bimodal and ranges from cloud-free views to total occlusion.}
    \label{fig:cloud_coverage_train}
\end{figure}

\begin{figure}[h!tb]
    \includegraphics[width=\linewidth]{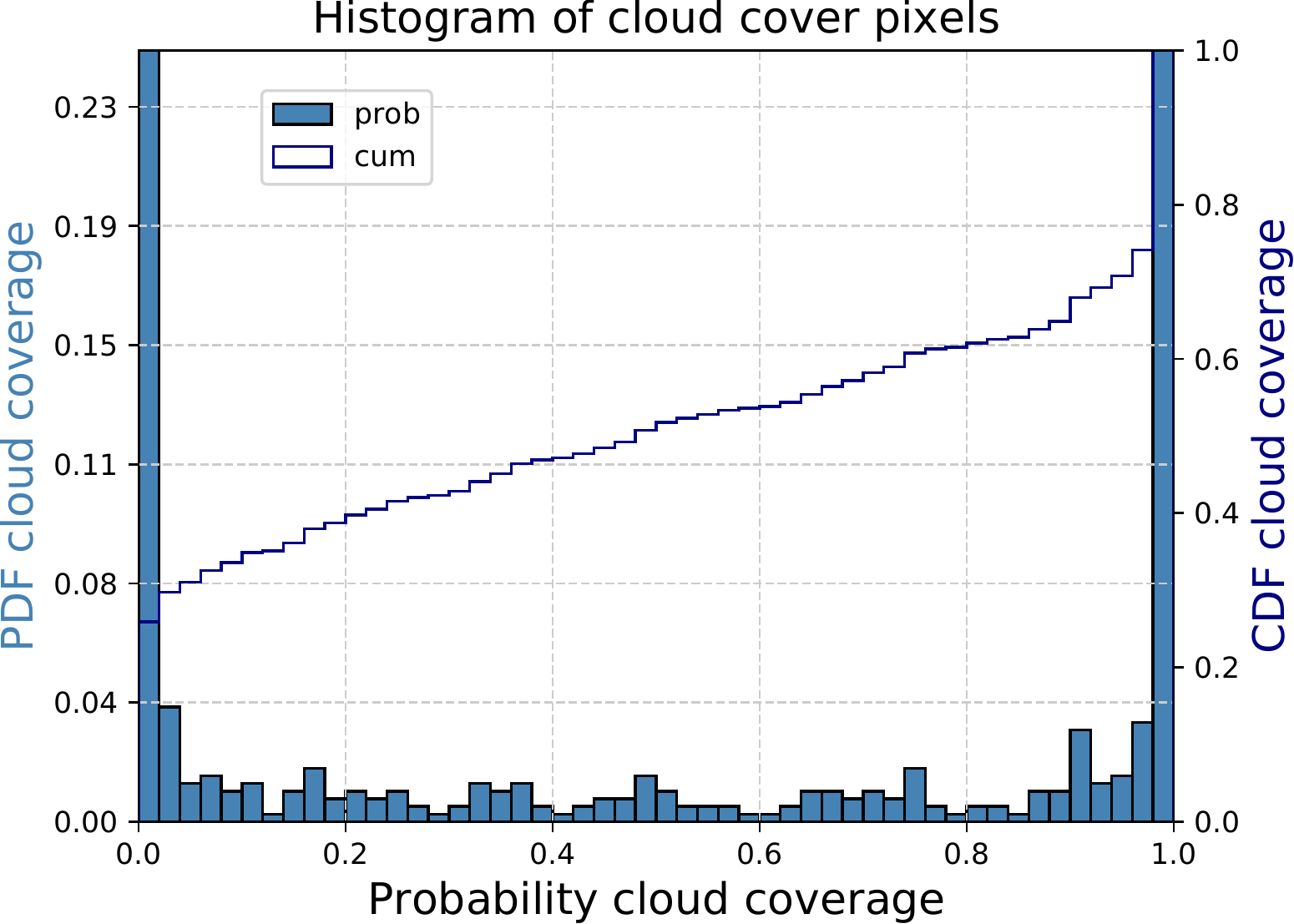}
    \caption{Statistics of cloud coverage of SEN12MS-CR-TS test split, computed on full-scene images via the detector of \cite{Zupanc}. On average, approximately $50 \%$ $(\pm 42 \%)$ of occlusion is observed. The empirical distribution of cloud coverage is bimodal and ranges from cloud-free views to total occlusion.}
    \label{fig:cloud_coverage_test}
\end{figure}

\subsection{Related Work} As the presence of clouds in optical satellite imagery poses a severe hindrance for remote sensing applications, there has been plenty of preceding research on cloud removal methods \cite{Enomoto_Sakurada_Wang_Fukui_Matsuoka_Nakamura_Kawaguchi_2017, Grohnfeldt_Schmitt_Zhu_2018, Singh_Komodakis_2018, sintarasirikulchai2018multi, Bermudez_Happ_Feitosa_Oliveira_2019, Rafique_Blanton_Jacobs, Sarukkai_Jain_Uzkent_Ermon_2019, meraner2020cloud, ebel2020multisensor}. The focus of this overview is on data sets for cloud removal methods. Much of the early work on cloud removal considered data of simulating cloudy observations \cite{Rafique_Blanton_Jacobs}. Copying cloudy pixel values from one image to another clear-view one \cite{Rafique_Blanton_Jacobs} captures the spectral properties of naturally cloudy observations more faithfully than synthetic noise (e.g. Perlin noise \cite{Perlin}) \cite{Enomoto_Sakurada_Wang_Fukui_Matsuoka_Nakamura_Kawaguchi_2017, Grohnfeldt_Schmitt_Zhu_2018, sintarasirikulchai2018multi}, but neither precisely reproduce the statistics of satellite images containing natural cloud occurrences \cite{ebel2020multisensor}. 
Consequently, our data set contain cloud-free as well as naturally occurring cloud-covered optical satellite recordings. The SEN12MS-CR data set \cite{ebel2020multisensor} provides a globally distributed collection of co-registered mono-temporal Sentinel-1 as well as cloudy and cloud-free Sentinel-2 observations. Our data set is an extension of SEN12MS-CR in the sense that we collect repeated measures per ROI and thereby provide a time-series of co-registered S1 and S2 observations, gathered such that matched observations of both modalities are no more than two weeks apart. In comparison to the preceding data set, ours allows integrating information not solely across different sensors, but also across different points in time distributed throughout the year. Similarly, the work of \cite{Sarukkai_Jain_Uzkent_Ermon_2019} allows for time-series cloud removal by providing a collection of tri-temporal RGB(NIR)-channel optical data and corresponding models. Our contribution extends this work by providing true multi-modal data recorded by two distinct sensors, synthetic aperture radar Sentinel-1 measurements as well as 13-bands multi-spectral Sentinel-2 observations. Furthermore, the length of each time series is increased considerably, from 3 to 30 samples. Finally, \cite{Sarukkai_Jain_Uzkent_Ermon_2019} exclude observations with greater than $30$ percent cloud coverage from their data set, which deviates from real conditions. Our approach aims to model the complete spectrum of cloud coverage, including conditions commonly encountered by remote sensing practitioners. In sum, our work and its main contribution, a large-scale multi-modal multi-temporal data set for cloud removal in optical satellite imagery, build on a history of research and improve upon the current state of image reconstruction in remote sensing by providing a novel, carefully curated data set.

\section{Data} \label{data}

\begin{figure*}[h!tb]
  \centering
  \begin{subfigure}[b]{0.49\linewidth}
    \includegraphics[width=\linewidth]{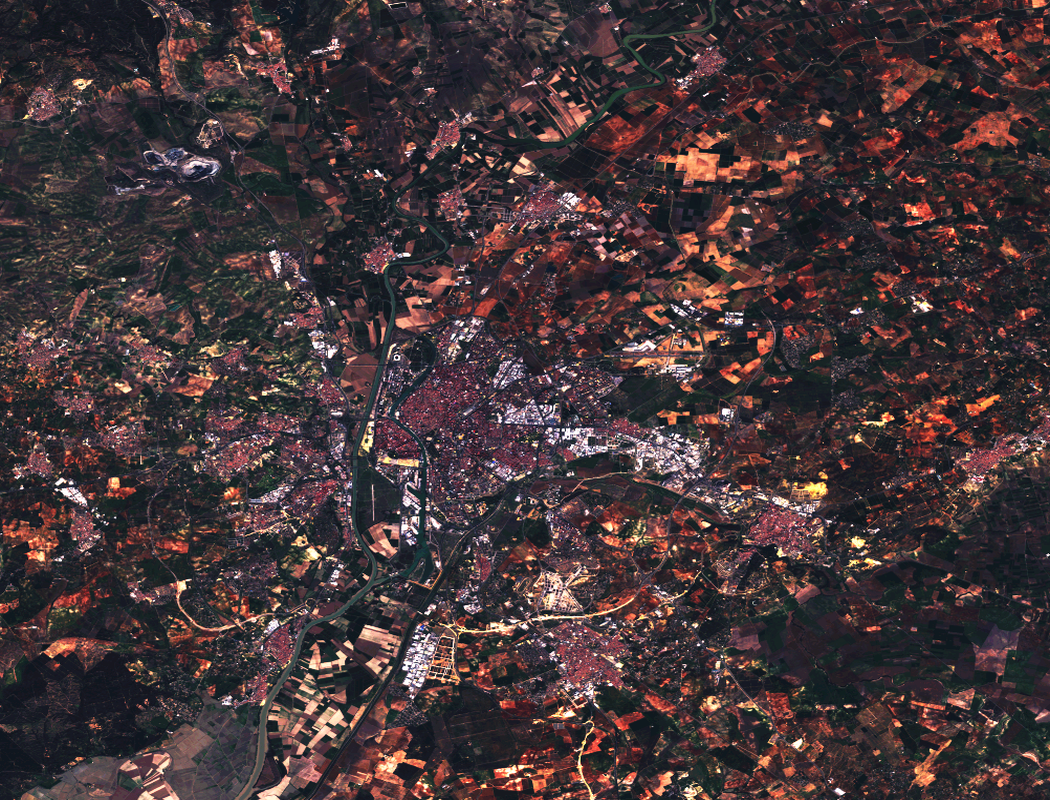}
  \end{subfigure}
    \begin{subfigure}[b]{0.49\linewidth}
    \includegraphics[width=\linewidth]{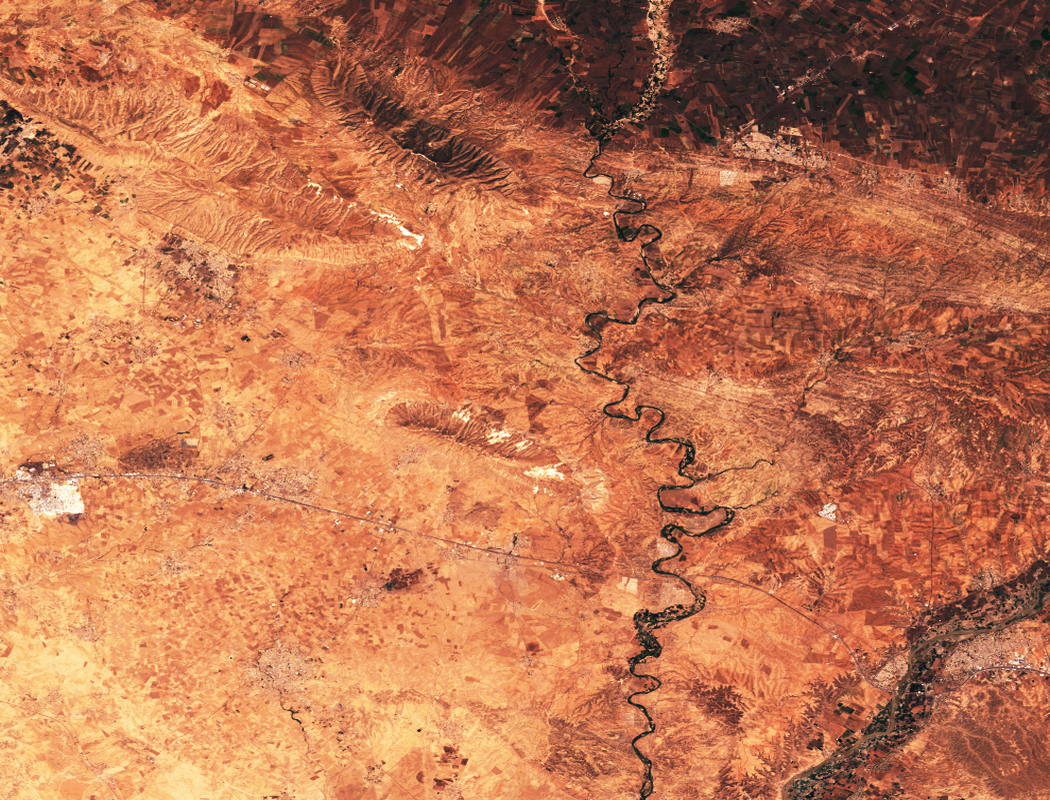}
  \end{subfigure}
  \label{fig:coffee}
  \begin{subfigure}[b]{0.49\linewidth}
    \includegraphics[width=\linewidth]{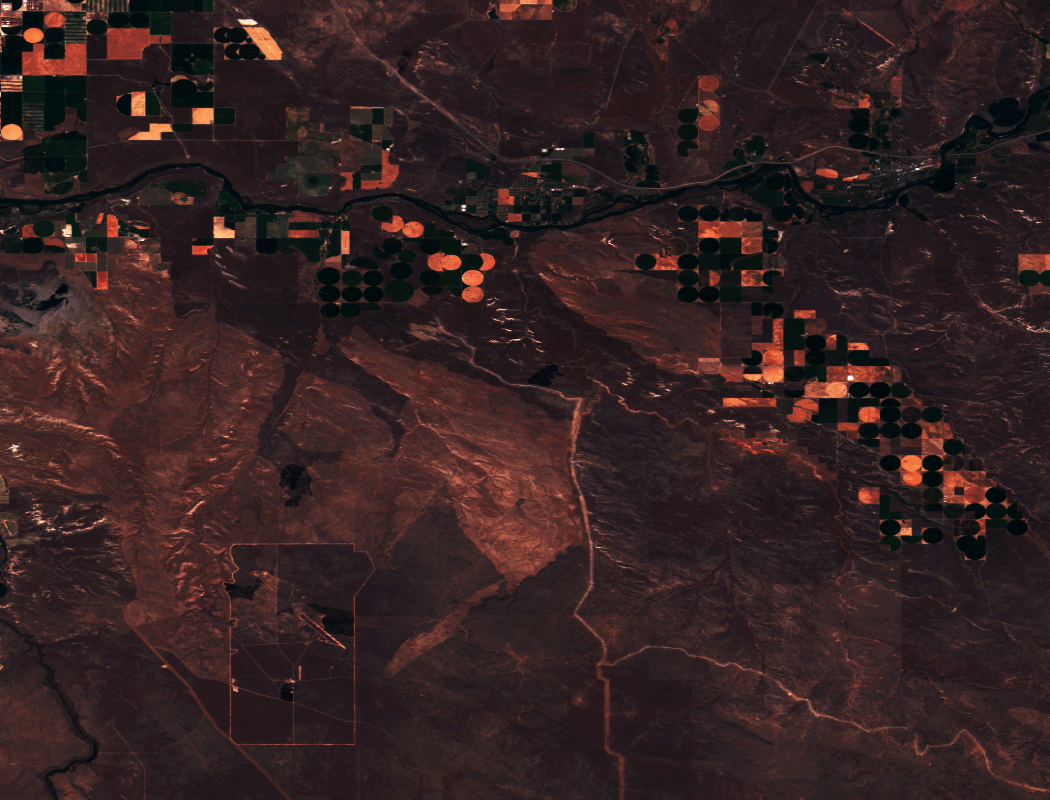}
  \end{subfigure}
    \begin{subfigure}[b]{0.49\linewidth}
    \includegraphics[width=\linewidth]{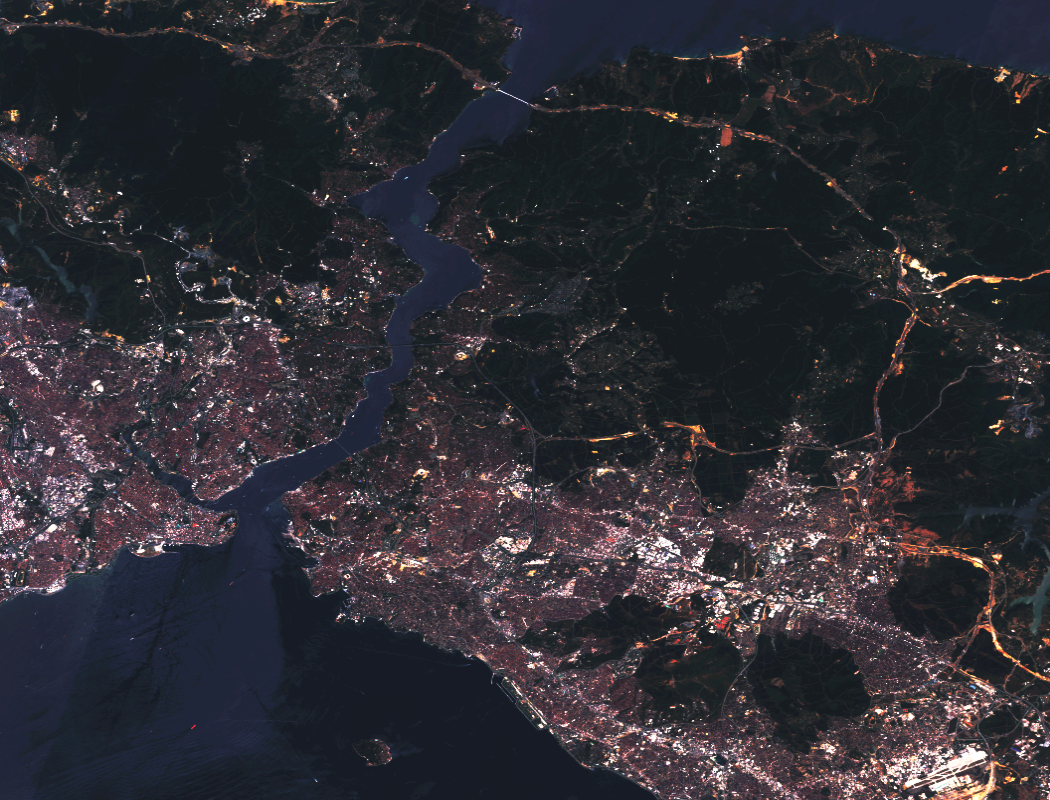}
  \end{subfigure}
  \caption{Four different regions contained in SEN12MS-CR-TS, highlighting the diversity of sampled landcovers. The depicted S2 observations (RGB channels) are cloud-free samples of their respective time series. The average ROI covers about $40 \times 40$ $km^2$ and is split into over 700 patch samples, each patch of size $256 \times 256 \: px^2$.}
  \label{fig:diverseROI}
\end{figure*}

This work introduces SEN12MS-CR-TS, a multi-modal and multi-temporal data set for training and evaluating global and all-season cloud removal methods. The data set consists of 53 globally distributed regions of interest, curated as detailed in section  \ref{curation}. The ROI are over $4000 \times 4000$ $px^2$ each, covering about $40 \times 40$ $km^2$ of land such that the total surface area covered by the data set is over $80,000$ $km^2$. Of all collected regions of interest, 40 are defined as a training split and 13 as a hold-out split to evaluate cloud removal approaches on. For every ROI we collect 30 co-registered and paired S1 and S2 full-scene images evenly spaced in time throughout the year of 2018. Each acquired image is inspected and quality-controlled manually. The cloud-free Sentinel-2 (RGB-channel) observations of four example ROI illustrating the diversity of our data set are illustrated in Fig. \ref{fig:diverseROI}. The spatial distribution of all ROI is depicted in Fig. \ref{fig:map} and highlights the global sampling of our data set. The empirical distribution of the cloud coverage of all optical observations is computed as detailed in section \ref{sub:cloudmask} and the statistics are presented in Fig. \ref{fig:cloud_coverage_train} and \ref{fig:cloud_coverage_test} for the train and the test splits, respectively. Importantly, the data set is curated without excluding any interval of cloud coverage such that the collected observations also reflect conditions of high cloud coverage as commonly encountered in practice \cite{King_Platnick_Menzel_Ackerman_Hubanks_2013}. The data is made available under \textit{\url{https://patrickTUM.github.io/cloud_removal}}. It is about 2 Tb in size and compatible with the SEN12MS-CR data set \cite{ebel2020multisensor}. That is, no train ROI of SEN12MS-CR are part of our data set's test ROI and vice versa.

\subsection{Data collection} \label{curation}

All curated data are recorded via the SAR Sentinel-1 and multi-spectral Sentinel-2 (level 1-C top-of-atmosphere reflectance product) instruments of ESA's Copernicus mission. The recorded observations are acquired via Google Earth Engine \cite{gorelick2017google} and a custom semi-automatic processing pipeline. We randomly sample the geospatial locations of $53$ regions of interest from SEN12MS-CR \cite{ebel2020multisensor}. To minimize mosaicing, observations of cells covered by a single pass are collected. The samples are referenced within the World Geodetic System 1984 (WGS84) coordinate system. For every ROI, 30 time intervals are evenly spaced throughout the year of 2018. For every time interval a co-registered, geo-referenced and full-scene S1 image as well as a paired full-scene S2 image (level 1-C) are collected. The acquisition within the same interval window is such that corresponding multi-modal images are no more than two weeks apart. Further statistics regarding the pairing of observations are provided in appendix \ref{appendix:a}.

\subsection{Preprocessing} \label{preprocessing}

To prepare the collected raw data and translate it into a format that neural networks for cloud removal can handle the following preprocessing steps are taken: Each band of every observation is upsampled to 10m resolution (i.e. to the native resolution of Sentinel-2's bands 2,3,4 and 8). Every full-scene image is sliced into non-overlapping patches of dimensions $256 \times 256 px^2$. The S1 observations are processed via the Sentinel-1 toolbox \cite{veci2014sentinel} (including border and thermal noise removal, radiometric calibration, orthorectification) and decibel-transformed. An example patch-wise tuple of paired S1 and S2 data is illustrated in Fig. \ref{fig:sample_patches_masks}. Input patches to any ResNet model \cite{he2016deep} are preprocessed in line with the pipeline of \cite{meraner2020cloud} as follows: the VV, VH channels of $S1$ observations are value-clipped in the ranges $[-25;0], [-32.5;0]$ and rescaled to the interval $[0;2]$, while $S2$ patches are value-clipped to $[0;10000]$ and normalized to the range $[0;5]$. For all other networks with a different backbone architecture, preprocessing is done as follows: each patch is value-clipped and then rescaled for every pixel to take normalized values within the unit range of $[0,1]$. The modalities $S1$ and $S2$ are value-clipped within the intervals of $[-25;0]$ and $[0;10000]$, respectively. This way, we follow the preprocessing protocol of the preceding work and avoid any unnecessary adjustments, for the sake of simplicity. For evaluation, the pixel values of all input patches, target images and predictions are re-mapped to the unit interval $[0,1]$, where the goodness of predictions is assessed according to the metrics stated in section \ref{sub:metrics}.

\subsection{Cloud detection \& mask computation} \label{sub:cloudmask}
In order to analyse the statistics of cloud coverage in SEN12MS-CR-TS and to model the spatio-temporal extent of clouds we compute binary cloud masks $m$. For each optical image, the masks $m$ are computed on-the-fly via the cloud detector of s2cloudless \cite{Zupanc}, which provides a binary mask of pixel-wise values in $\left \{ 0,1 \right \}$ that indicate cloud-free and cloud-covered pixels, respectively. The cloud mask accuracy of s2cloudless is reported to be on par with the multi-temporal classifier MAJA \cite{lonjou2016maccs}, but the considered detector can be applied on mono-temporal satellite observations. Note that, alternatively to s2cloudless, the masks $m$ may be computed via a dedicated neural network for cloud detection \cite{jeppesen2019cloud, lopez2021benchmarking}. However, s2cloudless has proven to be lightweight and provide sufficient performance at little extra computational cost in run time or memory, making it an appealing cloud detector to be applied on a large-scale date set such as SEN12MS-CR-TS. Example cloud detections are illustrated in Fig. \ref{fig:sample_patches_masks}.

\section{Methods} \label{methods}

We consider two distinctively different methods to highlight the benefits of our curated data set and the diverse tasks it allows to approach. The first method is a neural network reconstructing cloud covered pixels in time series of multi-modal data to predict a single target image acquired at a cloud-free time point. The second approach introduces a neural network that performs sequence-to-sequence cloud removal, that is, it predicts a time series of cloud-free observations the size of the cloudy input sequence.

\subsection{Multi-temporal multi-modal cloud removal}\label{method:seq2point}

\begin{figure*}[h!tb] 
    \includegraphics[width=\linewidth]{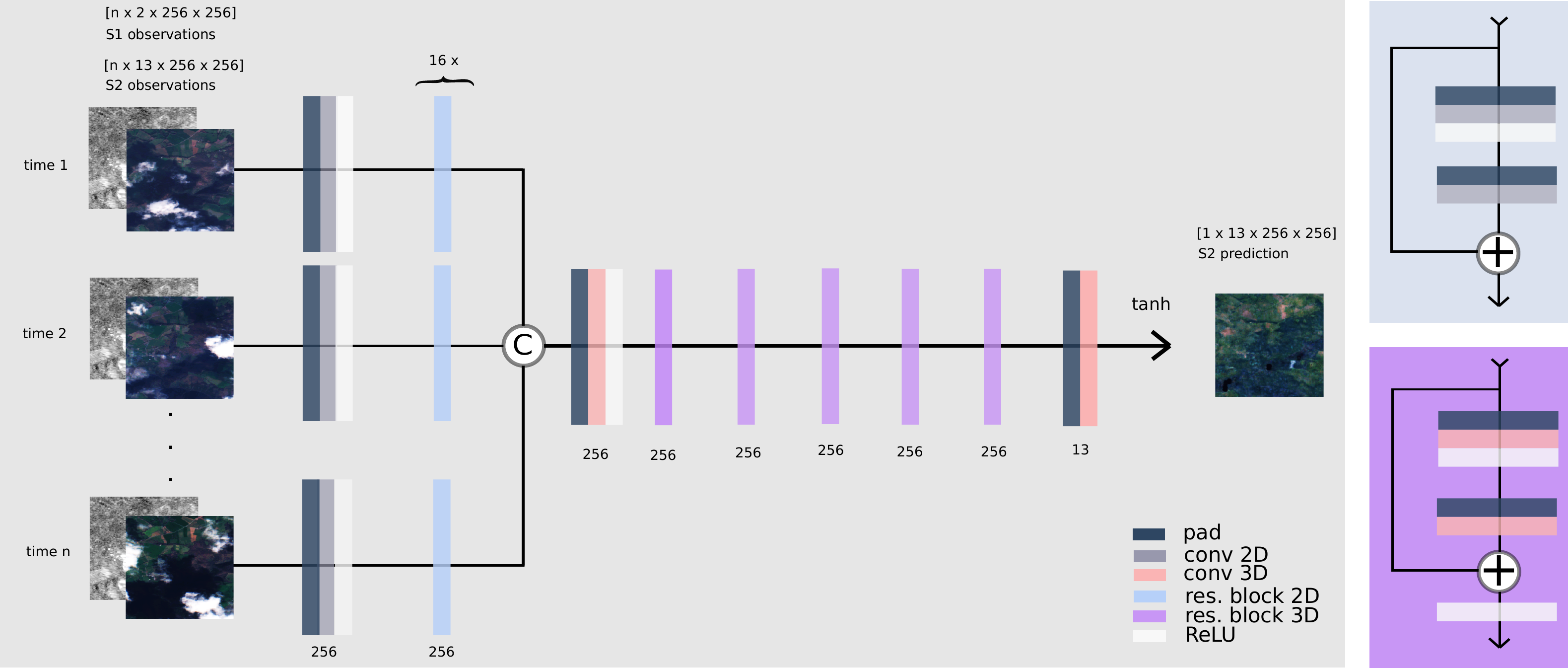}
    \caption{A conceptual illustration of the sequence-to-point cloud removal architecture $G_{seq2point}$. The network is based on the architecture of \cite{Sarukkai_Jain_Uzkent_Ermon_2019} and consists of $n$ siamese ResNet branches \cite{meraner2020cloud} doing single time point cloud removal on $n$ individual time points. Subsequently, the feature maps are stacked in the temporal dimension and 3D convolutions are applied to integrate information across time. The output of the network is a single cloud-free image prediction.}
    \label{fig:architecture_seq2point}
\end{figure*}

For multi-temporal multi-modal cloud removal we consider a deep neural network that builds on the generator of \cite{Sarukkai_Jain_Uzkent_Ermon_2019}. Our model receives a sequence of $t=1, ..., n$ input tuples $\left (S1, S2 \right )_t$ and predicts a cloud-removed multispectral image $\hat{S2}$. The architecture of the proposed model uses a ResNet \cite{he2016deep} backbone, with Siamese residual branches processing the individual time points until their information gets integrated. That is, we replaced the pairwise concatenation of 2D feature maps in \cite{Sarukkai_Jain_Uzkent_Ermon_2019} by stacking features in the temporal domain, followed by 3D convolutions. Moreover, as the first part of the generator of \cite{Sarukkai_Jain_Uzkent_Ermon_2019} is effectively a single time point cloud removal sub-network (as each time point is processed individually up to this point), we substitute this component by the established ResNet-based \cite{he2016deep} cloud removal network of \cite{meraner2020cloud}. Subsequently, the feature maps are stacked in the temporal dimension and 3D convolutions are applied to integrate information across time. The output of the network is a single cloud-free image prediction $\hat{S2}$. A schematic overview of the described architecture is shown in Fig. \ref{fig:architecture_seq2point}.

\subsection{Internal Learning for sequence-to-sequence cloud removal} \label{subsec:seq2se2method}

The sequence-to-sequence cloud removal method \cite{ebel2021seq2seq} follows the 3D Encoder-Decoder architecture of \cite{zhang2019internal}, constituted of an encoder as well as a decoder component. Both components are arranged symmetrically in the style of U-Net \cite{ronneberger2015u} and linked via skip connections between paired layers. The input to the network is a sequence of multi-temporal $S1$ samples and its output is a sequence of multi-temporal cloud-removed $S2$ predictions. With regards to its input-to-output mapping, the proposed architecture resembles earlier SAR-to-optical translation method  \cite{Fuentes_Reyes_Auer_Merkle_Henry_Schmitt_2019, wang2019sar}. Similar to these earlier domain translation approaches, our network learns information of the target domain (i.e. the optical imagery) via the supervision signal. Different from these approaches, the internal learning framework described below removes clouds and directly learns to denoise the target image sequence.

\begin{figure*}[h!tb] 
    \includegraphics[width=\linewidth]{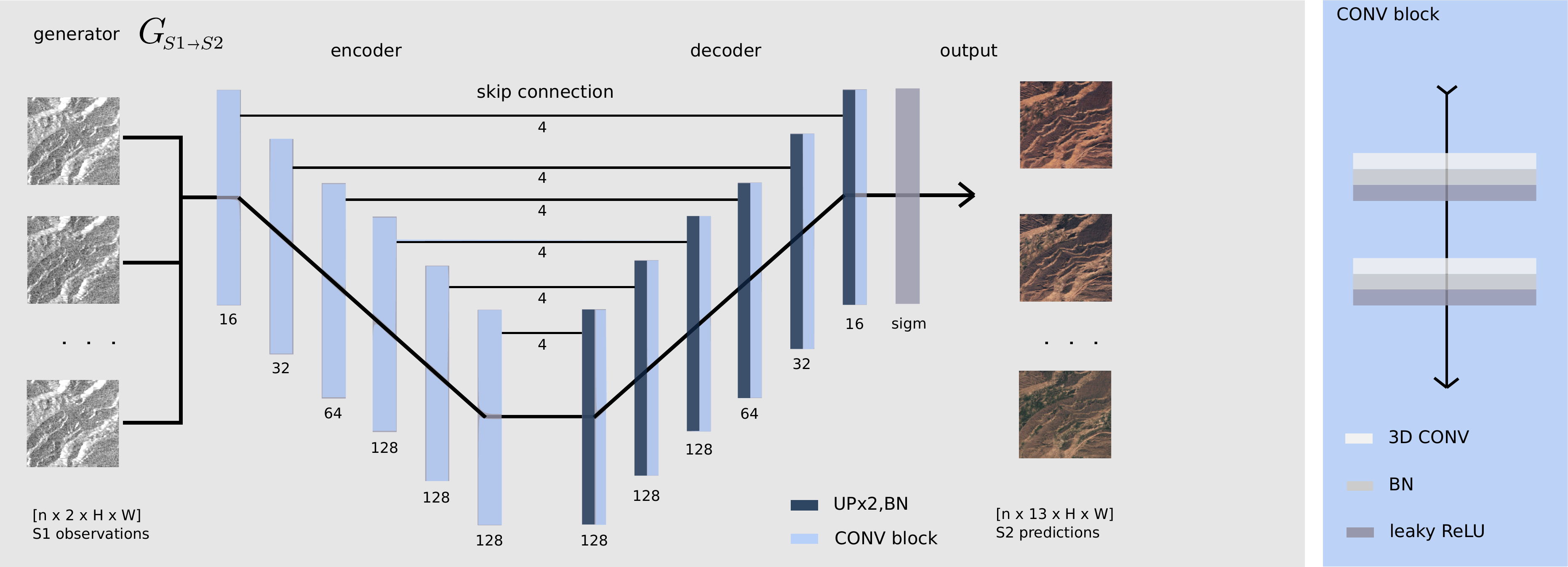}
    \caption{A conceptual illustration of the 3D Encoder-Decoder architecture $G_{seq2seq}$ employed in the sequence-to-sequence cloud removal model \cite{ebel2021seq2seq}. The network is based on the architecture of \cite{zhang2019internal} and consists of encoder and decoder parts arranged symmetrically in the style of U-Net \cite{ronneberger2015u}, with skip connections between paired layers. Input to the network is a batch of multi-temporal $S1$ observations. The output is a predicted batch of multi-temporal multi-spectral $S2$ observations. For the ablation model considered in section \ref{exp:seq2seq}, Gaussian noise is used as an input as in \cite{ulyanov2018deep, zhang2019internal}.}
    \label{fig:architecture_seq2seq}
\end{figure*}

The architecture of the network is summarized in Fig. \ref{fig:architecture_seq2seq}. Note, that the key difference between the given model and the sequence-to-point method of section \ref{method:seq2point} (depicted in Fig. \ref{fig:architecture_seq2point}) is in the output dimensions: Whereas the sequence-to-point architecture maps a sequence of $n$ cloudy inputs to a single cloud-removed prediction, the sequence-to-sequence approach preserves the temporal information by mapping to a time series of $n$ cloud removed outputs. Moreover, the point estimator receives tuples of $S1$ and $S2$ inputs, whereas the network of Fig. \ref{fig:architecture_seq2seq} is driven solely by $S1$ data (or Gaussian noise, as proposed in \cite{ulyanov2018deep, zhang2019internal}). Finally, the sequence-to-point network of Fig. \ref{fig:architecture_seq2point} builds on the Siamese architecture of \cite{Sarukkai_Jain_Uzkent_Ermon_2019} with a ResNet backbone \cite{meraner2020cloud} plus 3D convolutions, whereas the sequence-to-sequence approach of Fig. \ref{fig:architecture_seq2seq} follows a 3D convolutional variant of U-Net \cite{ronneberger2015u}, as proposed in \cite{zhang2019internal}.

The training procedure of the sequence-to-sequence network follows that of internal learning for image inpainting \cite{ulyanov2018deep, zhang2019internal}, which is formalized in Algorithm \ref{algo:internal}. In this framework, for a given target sequence, a neural network is trained from scratch directly on the target sequence (without any need for additional or cloud-free training data) in order to reconstruct its noisy pixels. The observations exhibit spatio-temporal regularities and patterns (i.e. signal in the data), which is first modeled and learned by the network. The irregularities in the sequence (i.e. noise in the target data) are only internalized after, similar to a conventionally-trained network overfitting to noise on training data. The internal learning approach exploits this signal-noise dichotomy and teaches a model to reconstruct cloud-covered pixels in the target sequence of $S2$ observations, without need for any external or cloud-free training data. In detail, a neural network is initialized and trained from scratch directly on the target sequence. At each iteration, the model receives input driving its activations (e.g. Gaussian noise or $S1$ recordings) and predicts a sequence $\hat{S2}$. The predictions $\hat{S2}$ are compared against the target sequence $S2$ (e.g. according to a cost function ${L}_{all}$ as in eq. \ref{eq:Lall_seq2seq}) and the network learns to reproduce the cloud-free pixels. The training stops before the network overfits to internalizing the cloudy pixels. 

	\begin{algorithm}
		\caption{Internal Learning to Remove Clouds}
		\begin{algorithmic}[1]
			\Procedure{Seq2SeqDeclouding}{$S1, S2, iterMax$}
			\State $G_{S1 \rightarrow S2}$ = init. new NeuralNetwork()
			\State iterCount = $0$

			\While {iterCount $<$ iterMax}
			\State $\hat{S2} = G_{S1 \rightarrow S2}(S1)$
			\State $G_{S1 \rightarrow S2}.backpropagate(\mathcal{L}_{all}(S2, \hat{S2}))$
            \State iterCount = iterCount + 1
			\EndWhile
			\State Return $\hat{S2}$
			\EndProcedure
		\end{algorithmic}
	  \label{algo:internal}
	\end{algorithm}

With respect to its application and functionality, our sequence-to-sequence neural network resembles classical low-rank \& sparse signal decomposition methods \cite{de2001robust, donoho2006compressed, zhu2010tomographic, zhu2012sparse}: First, while neural networks are typically trained on a dedicated training data set separated from the test observations, numeral signal decomposition methods can be directly utilized on the data of interest. Similarly, our model can be directly applied on the test data. Second, unmixing of signals is very generic and can be applied to matrices as well as tensors. In comparison, the deep image prior approach applies to single images as well as time series \cite{ulyanov2018deep, zhang2019internal}, too. Finally, the decomposition itself is into a low-rank part and a sparse component. The low-rank part denotes the data‘s compact representation and regularities. That is, spatial, spectral or temporal (auto-)correlations such as the land cover mapped by a satellite. The sparse component consists of the irregular part of the data which has only a few non-zero entries, such as the appearance of clouds. In comparable terms, the internal learning technique allows our network to discover the regularities in the data and generalizing it to cloud-covered samples, before overfitting to the noise.

\section{Experiments and Results} \label{experiments}

This methods details the experimental design and the corresponding results on the considered cloud removal methods as well as their ablation variants. 
Section \ref{sub:metrics} specifies the measures of goodness used to assess the quality of the individual techniques' predictions. Section \ref{sub:baselines} introduces the baselines compared against the proposed model of \ref{method:seq2point} on the sequence-to-point cloud removal task. Sections \ref{exp:seq2point} and \ref{exp:seq2seq} detail the experiments and outcomes for the sequence-to-point and sequence-to-sequence cloud removal tasks, respectively.

\subsection{Metrics} \label{sub:metrics}

We evaluate quantitative performance in terms of 
normalized root mean squares error (NRMSE), Peak Signal-to-Noise Ratio (PSNR), structural similarity (SSIM) \cite{Wang_Bovik_Sheikh_Simoncelli_2004} and Spectral Angle Mapper (SAM) \cite{kruse1993spectral}, defined as
$$NRMSE(x,y) = \sqrt{\frac{1}{C \cdot H \cdot W} \sum_{c=h=w=1}^{C, H, W} (x_{c, h, w}-y_{c, h, w})^2}$$ 
$$PSNR(x,y) = 20 \cdot log_{10} \left( \frac{1}{NRMSE(x,y)} \right)$$ 
$$SSIM(x,y) = \frac{(2 \mu_x\mu_y + \epsilon_1)(2 \sigma_{xy} + \epsilon_2)}{(\mu_x+\mu_y+\epsilon_1)(\sigma_x + \sigma_y + \epsilon_2)}$$
\small $$SAM(x,y) = cos^{-1} \left( \frac{\sum^{C, H, W}_{c=h=w=1} x_{c, h, w} \cdot y_{c, h, w}}{\sqrt{\sum^{C, H, W}_{c=h=w=1} x_{c, h, w}^2 \cdot \sum^{C, H, W}_{c=h=w=1} y_{c, h, w}^2}} \right)$$
\normalsize
with images $x,y$ compared via their respective pixel-values $x_{c, h, w}, y_{c, h, w} \in [0,1]$, dimensions $C=3$, $H=W=256$, means $\mu_x, \mu_y$, standard deviations $\sigma_x, \sigma_y$, covariance  $\sigma_{xy}$ as well as constants  $\epsilon_1$, $\epsilon_2$ to stabilize the computation. 
NRMSE belongs to the class of pixel-level metrics and quantifies the average discrepancy between target and predicted pixels in 
units of the measure of interest. PSNR is evaluated on the whole image and quantifies the signal-to-noise ratio of the prediction as a reconstruction of the target image. SSIM is another image-wise measure that builds on PSNR and captures the structural similarity of the prediction to the target in terms of perceived change, contrast and luminance \cite{Wang_Bovik_Sheikh_Simoncelli_2004}. The SAM measure is a third image-level metric that provides the spectral angle between the bands of two multi-channel images \cite{kruse1993spectral}. For further analysis, the pixelwise NRMSE is evaluated in three manners: 1) over all pixels of the target image (as per convention), 2) only over cloud-covered pixels (visible in neither of any input optical sample) to measure reconstruction of noisy information, as well as 3) only over cloud-free pixels (visible in at least one input optical patch) quantifying preservation of information. The pixel-wise masking is performed according to the cloud mask given by the detector of \cite{Zupanc}.

\subsection{Baseline Methods} \label{sub:baselines}

To put the performances of our proposed model and ablations into context, we consider the following baseline methods. First ("least cloudy"), taking the least-cloudy input observation and forwarding it without further modification to be compared against the cloud-free target image. This provides a measure of how hard the cloud removal task is with respect to the extent of cloud-coverage present in the data. Second ("mosaicing"), we perform a mosaicing method that averages the values of pixels across cloud-free time points, thereby integrating information across time. That is, for any pixel, if there is a single clear-view time point then its value is copied, for multiple cloud-free samples the mean is formed and in case no cloud-free time point exists, then a value of 0.5 is taken as a proxy. This is to avoid any extreme values, such as cloudy pixels of high-intensity. The mosaicing technique provides a measure of how much information can be reconstructed across time, from multi-spectral optical observations exclusively. Third, ResNet refers to a residual neural network as described and trained in sections \ref{method:seq2point} as well as \ref{exp:seq2point}. The architecture is based on the model of \cite{meraner2020cloud}, and serves as a relevant baseline because parts of this model are used as Siamese residual branches within our model, as detailed in section \ref{method:seq2point}. It provides an estimate of how well a point-to-point cloud removal model can perform as a baseline. Fourth, the baseline STGAN denotes the "Branched ResNet generator (IR)" architecture of \cite{Sarukkai_Jain_Uzkent_Ermon_2019}. It is a sequence-to-point cloud removal model, and the architecture of our own sequence-to-point neural network closely follows its design, as detailed in section \label{method:seq2point}.
In sum, the purpose of assessing these baselines is to analyze whether trivial solutions to the multi-modal multi-temporal sequence-to-point cloud removal problem exist, and how any more sophisticated deep learning approach compares against these methods and our proposed model trained on SEN12MS-CR-TS.

\subsection{Sequence-to-Point Cloud Removal} \label{exp:seq2point}

This section details the training specifics of the sequence-to-point cloud removal architecture introduced in section \ref{method:seq2point}.
As detailed in section \ref{method:seq2point}, up to the temporal concatenation layer, we use a version of the ResNet-based \cite{he2016deep} cloud removal network of \cite{meraner2020cloud} and pre-trained it on SEN12MS-CR \cite{ebel2020multisensor} according to the training specifics of \cite{meraner2020cloud}. All our considered sequence-to-point cloud removal networks and ablation models share this pre-trained single-temporal cloud removal network as a starting point for the sake of comparability and in order to reduce the duration of training. The networks are trained for a total of 10 epochs on 1 tuple of patches per location for every ROI in the training split. For training, the input $S2$ patches are filtered to display within $0$ to $50$ percent of cloud coverage. The target $S2$ patch is selected to be the sample showing the minimum cloud coverage over the given time series, i.e. it is not necessarily temporally preceding or following the input patches. For the first 25000 steps in the training procedure, the networks are trained with the initial ResNet Siamese components frozen, exclusively optimizing the subsequent 3D-convolution layers. After the steps with the pre-trained weights frozen and once the deeper layers have been calibrated to the initial network's latent feature maps, the full network is trained end-to-end for the remainder of the process. During training, the network minimizes the loss $\mathcal{L}_\text{all}$ 

\begin{equation}
\label{eq:Lall_seq2point}
\mathcal{L}_\text{all} = \lambda_{L1} \mathcal{L}_{L1} + \lambda_{perc} \mathcal{L}_{perc}
\end{equation}

\begin{equation}
\label{eq:L1_seq2point}
\mathcal{L}_\text{L1} = ||S2-\hat{S2}||_1
\end{equation}

\begin{equation}
\label{eq:Lperc_seq2point}
\mathcal{L}_{perc} = ||VGG16(S2), VGG16(\hat{S2})||_2
\end{equation}

with $\lambda_{L1} = 100$ according to \cite{Sarukkai_Jain_Uzkent_Ermon_2019} and $\lambda_{perc} = 1$ as hyperparameters weighting the individual pixel-wise loss $\mathcal{L}_\text{L1}$ and the perceptual loss $\mathcal{L}_{perc}$. The perceptual loss is computed by means of an auxiliary VGG16 network \cite{Simonyan_Zisserman_2014} resulting in sharper image reconstructions \cite{johnson2016perceptual}. In comparison to other VGG16 pre-trained on classical computer vision data sets such as ImageNet \cite{russakovsky2015imagenet} and thus limited to RGB channel data, we pre-trained a VGG16 for landcover classification on the SEN12MS data set \cite{schmitt2019sen12ms} according to the training protocol of \cite{schmitt2021remote}. The proposed sequence-to-point cloud removal network (and its ablation variants) are optimized via ADAM \cite{kingma2014adam}, with a learning rate of $0.0002$ and momentum parameters $\left [ 0.5, 0.999 \right ]$ as in \cite{Sarukkai_Jain_Uzkent_Ermon_2019}. A batch size of $1$ tuple of samples per iteration is used for training.

\begin{figure*}[h!tb]
  \centering
  \begin{subfigure}[b]{0.2\linewidth}
    \includegraphics[width=\linewidth]{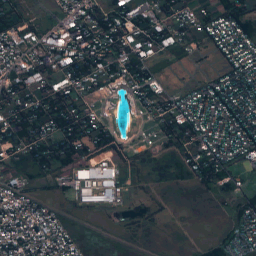}
  \end{subfigure}
  \begin{subfigure}[b]{0.2\linewidth}
    \includegraphics[width=\linewidth]{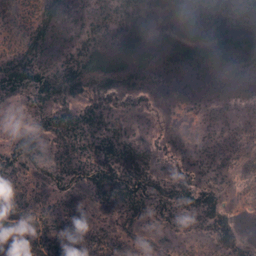}
  \end{subfigure}   
    \begin{subfigure}[b]{0.2\linewidth}
    \includegraphics[width=\linewidth]{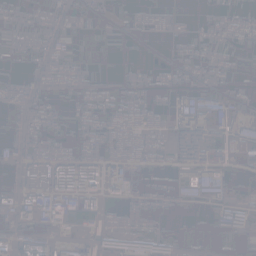}
  \end{subfigure}
      \begin{subfigure}[b]{0.2\linewidth}
    \includegraphics[width=\linewidth]{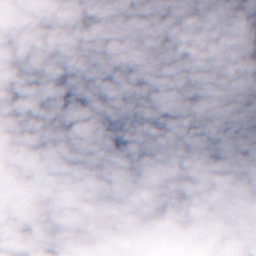}
  \end{subfigure}
  \begin{subfigure}[b]{0.2\linewidth}
    \includegraphics[width=\linewidth]{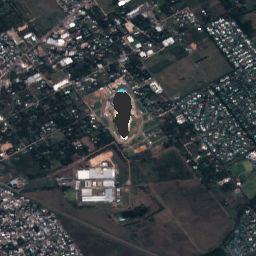}
  \end{subfigure}
    \begin{subfigure}[b]{0.2\linewidth}
    \includegraphics[width=\linewidth]{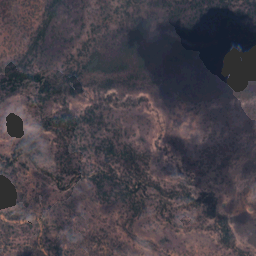}
  \end{subfigure}
    \begin{subfigure}[b]{0.2\linewidth}
    \includegraphics[width=\linewidth]{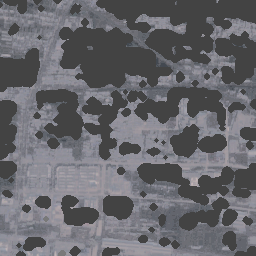}
  \end{subfigure}
  \begin{subfigure}[b]{0.2\linewidth}
    \includegraphics[width=\linewidth]{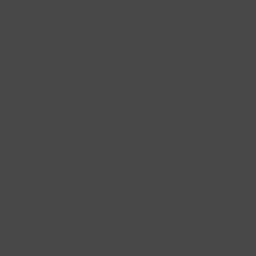}
  \end{subfigure}   
    \begin{subfigure}[b]{0.2\linewidth}
    \includegraphics[width=\linewidth]{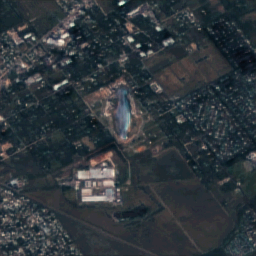}
  \end{subfigure}
      \begin{subfigure}[b]{0.2\linewidth}
    \includegraphics[width=\linewidth]{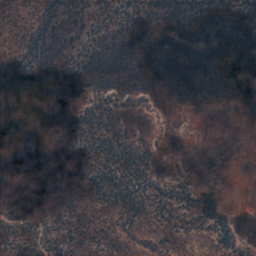}
  \end{subfigure}
  \begin{subfigure}[b]{0.2\linewidth}
    \includegraphics[width=\linewidth]{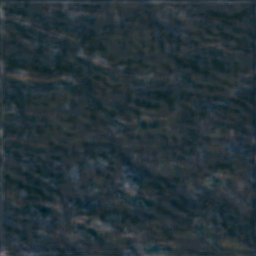}
  \end{subfigure}
    \begin{subfigure}[b]{0.2\linewidth}
    \includegraphics[width=\linewidth]{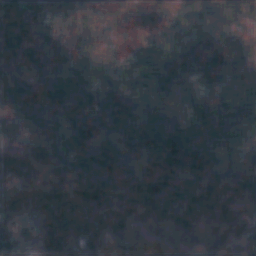}
  \end{subfigure}
    \begin{subfigure}[b]{0.2\linewidth}
    \includegraphics[width=\linewidth]{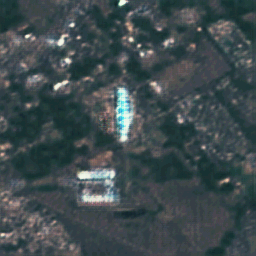}
  \end{subfigure}
  \begin{subfigure}[b]{0.2\linewidth}
    \includegraphics[width=\linewidth]{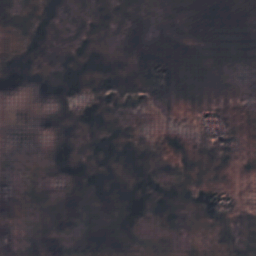}
  \end{subfigure}   
    \begin{subfigure}[b]{0.2\linewidth}
    \includegraphics[width=\linewidth]{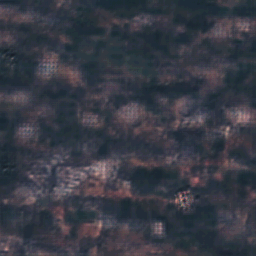}
  \end{subfigure}
      \begin{subfigure}[b]{0.2\linewidth}
    \includegraphics[width=\linewidth]{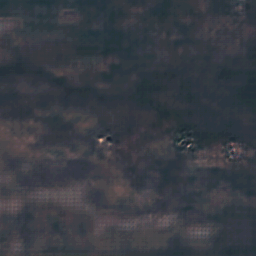}
  \end{subfigure}
  \begin{subfigure}[b]{0.2\linewidth}
    \includegraphics[width=\linewidth]{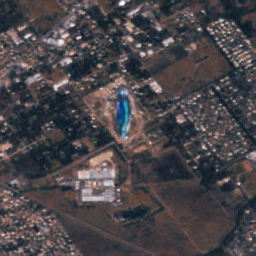}
  \end{subfigure}
    \begin{subfigure}[b]{0.2\linewidth}
    \includegraphics[width=\linewidth]{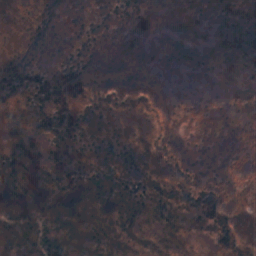}
  \end{subfigure}
    \begin{subfigure}[b]{0.2\linewidth}
    \includegraphics[width=\linewidth]{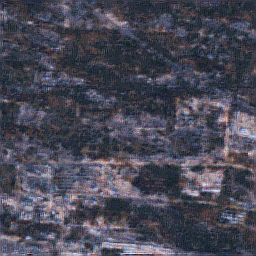}
  \end{subfigure}
  \begin{subfigure}[b]{0.2\linewidth}
    \includegraphics[width=\linewidth]{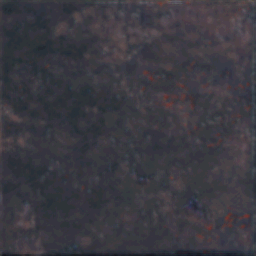}
  \end{subfigure}   
    \begin{subfigure}[b]{0.2\linewidth}
    \includegraphics[width=\linewidth]{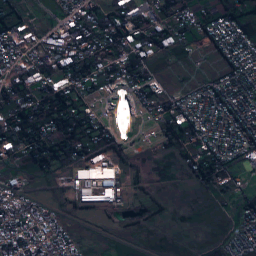}
  \end{subfigure}
      \begin{subfigure}[b]{0.2\linewidth}
    \includegraphics[width=\linewidth]{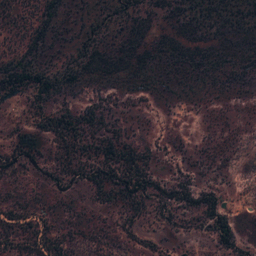}
  \end{subfigure}
  \begin{subfigure}[b]{0.2\linewidth}
    \includegraphics[width=\linewidth]{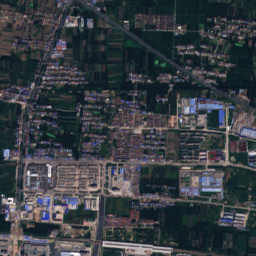}
  \end{subfigure}
    \begin{subfigure}[b]{0.2\linewidth}
    \includegraphics[width=\linewidth]{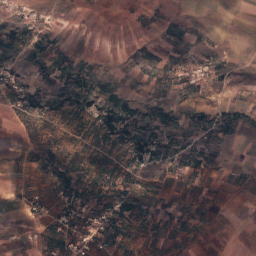}
  \end{subfigure}
  \caption{Exemplary predictions and cloud-free target images for all baselines reported in Table \ref{tab:baselines}. Columns: Four different samples from the test split. The considered cases are cloud-free, partly-cloudy, cloud-covered with no visibility except for a single time point and cloud-covered with no visibility in any time point. Rows: Predictions of least cloudy, mosaicing, ResNet, STGAN, ours (n=3) as well as the cloud-free reference image. The results show that the considered models outperform the simple heuristics. Moreover, the illustrations underline that multi-temporal and multi-modal data may benefit image reconstruction.}
  \label{fig:comparison_seq2point}
\end{figure*}

To evaluate performances on the test split, samples containing $S2$ observations from the complete range of cloud coverage (between $0$ and $100 \%$) are considered for input. Table \ref{tab:baselines} compared the results of our proposed model with the baselines detailed in section \ref{sub:baselines}. The results show that the proposed network outperforms the baselines in the majority of metrics, except for PSNR (where mosaicing comes first) and the NRMSE (clear) preservation metric (where the "least cloudy" approach performs best).  This demonstrates that a deep neural network approach can typically outperform trivial solutions to the multi-modal multi-temporal cloud removal problem. Exemplary outcomes for the considered baselines on four different samples from the test split are presented in Fig. \ref{fig:comparison_seq2point}. The considered cases are cloud-free, partly-cloudy, cloud-covered with no visibility except for a single time point and cloud-coverage with no visibility at any time point. The results show that the considered models typically outperform the simple heuristics. One exceptional case is least cloudy in the absence of clouds, which manages to accomplish a faithful prediction in such settings. Moreover, the illustrations underline that multi-temporal and multi-modal data may benefit image reconstruction: While most methods perform well in the cloud-free or partly-cloudy cases, multi-source integration is needed if individual time points contain dense cloud coverage over wide areas. When all input data is covered by thick clouds, then this poses a severe challenge for all approaches considered.
To analyze the benefits of including $S1$ SAR data, we perform an ablation study and compare a multi-sensor model against one only utilizing multi-spectral $S2$ input.
Table \ref{tab:sar_ablation} compared the results of the multi-modal model with an ablation version not using $S1$ SAR data. The comparison illustrates the benefits of including SAR data when reconstructing cloud-covered pixels. Next, we conduct an ablation experiment to assess the additional benefits of utilizing the introduced perceptual loss. Table \ref{tab:perceptual_Ablation} compared the results of our proposed model with an ablation version not using the perceptual loss (i.e. setting $\lambda_{perc} =0$ in eq \ref{eq:Lall_seq2point}). The outcomes imply that the usage of a perceptual loss results in cloud-removed predictions of a higher quality. Finally, we consider the extension of the proposed model into networks integrating 4 and 5 time points of input information. Table \ref{tab:more_time} compared the performance of our model as a function of input time points ($n=3,4,5$). The results indicate that considering longer time series may provide further improvements in terms of reconstructing cloud-covered information. In a final experiment on sequence-to-point cloud removal, Table \ref{tab:cloudy_percentage} reports the performance of our proposed model(n=3, with S1 and perceptual loss) as a function of cloud coverage. That is, for a given interval of cloud coverage, all $n=3$ input images are sampled to contain a corresponding extent of clouds. The outcomes show that image reconstruction performance is highly dependant on the percentage of cloud coverage. While performance decrease is not strictly monotonous with an increase in cloud coverage, a strong association persists.

\begin{table}[]
\scalebox{0.77}{
\begin{tabular}{@{}lllllllll@{}}
\toprule
model        & NRMSE (all) & NRMSE (cloudy) & NRMSE (clear) & PSNR & SSIM & SAM  \\ \midrule
least cloudy & 0.079        & 0.082          & \textbf{0.031}          & --- & 0.815 & 0.213 \\
mosaicing    & 0.062        & 0.064           & 0.036          & \textbf{31.68} & 0.811 & 0.250 \\
ResNet    & 0.060        & 0.062           & 0.040          & 26.04 & 0.810 & 0.212 \\
STGAN    & 0.057        & 0.059           & 0.050          & 25.42 & 0.818 & 0.219 \\
ours (n=3)   & \textbf{0.051}        & \textbf{0.052}           & 0.040          & 26.68 & \textbf{0.836} & \textbf{0.186} \\ \bottomrule
\end{tabular}
}
\caption{Quantitative evaluation of the proposed sequence-to-point model with baseline approaches in terms of normalized root mean squared error (NRSME), peak signal-to-noise ratio (PSNR), structural similarity  \cite{Wang_Bovik_Sheikh_Simoncelli_2004}
    (SSIM) and the Spectral Angle Mapper (SAM) \cite{kruse1993spectral} metric. Our model performs best in the majority of metrics, demonstrating that a deep neural network approach yields additional benefits over trivial solutions to the multi-modal multi-temporal cloud removal problem.
    }
    \label{tab:baselines}
\end{table}

\begin{table}[]
\scalebox{0.74}{
\begin{tabular}{@{}lllllll@{}}
\toprule
model          & NRMSE (all) & NRMSE (cloudy) & NRMSE (clear) & PSNR & SSIM & SAM  \\ \midrule
ours (no S1)   & 0.054        & 0.057           & 0.054          & 25.35 & 0.832 & 0.194 \\
ours (with S1) & \textbf{0.051}        & \textbf{0.052}           & \textbf{0.040}          & \textbf{26.68} & \textbf{0.836} & \textbf{0.186} \\ \bottomrule
\end{tabular}
}
\caption{Comparison of the proposed sequence-to-point model including SAR observations versus an ablation version without SAR observations in terms of normalized root mean squared error (NRSME), peak signal-to-noise ratio (PSNR), structural similarity \cite{Wang_Bovik_Sheikh_Simoncelli_2004}
    (SSIM) and the Spectral Angle Mapper (SAM) \cite{kruse1993spectral} metric. The comparison illustrates the benefits of including SAR data when reconstructing cloud-covered pixels.}
    \label{tab:sar_ablation}
\end{table}

\begin{table}[]
\scalebox{0.7}{
\begin{tabular}{@{}lllllll@{}}
\toprule
model          & NRMSE (all) & NRMSE (cloudy) & NRMSE (clear) & PSNR & SSIM & SAM  \\ \midrule
ours (no percept.)   & 0.052        & 0.053           & \textbf{0.039}          & 26.66 & 0.835 & \textbf{0.180} \\
ours (with percept.) & \textbf{0.051}        & \textbf{0.052}           & 0.040          & \textbf{26.68} & \textbf{0.836} & 0.186 \\ \bottomrule
\end{tabular}
}
\caption{Comparison of the proposed sequence-to-point model including perceptual loss versus an ablation version without perceptual loss in terms of normalized root mean squared error (NRSME), peak signal-to-noise ratio (PSNR), structural similarity \cite{Wang_Bovik_Sheikh_Simoncelli_2004}
    (SSIM) and the Spectral Angle Mapper (SAM) \cite{kruse1993spectral} metric. The outcomes imply that the usage of a perceptual loss during training results in cloud-removed predictions of a higher quality at test time.}
    \label{tab:perceptual_Ablation}
\end{table}

\begin{table}[]
\scalebox{0.77}{
\begin{tabular}{@{}lllllll@{}}
\toprule
model            & NRMSE (all) & NRMSE (cloudy) & NRMSE (clear) & PSNR & SSIM & SAM  \\ \midrule
ours (n=3) & 0.051        & 0.052           & 0.040          & 26.68 & 0.836 & 0.186 \\
ours (n=4) & 0.049        & 0.050           & 0.041               & \textbf{27.10} & 0.845 & \textbf{0.172} \\
ours (n=5) & \textbf{0.048}        & \textbf{0.048}           & \textbf{0.032}              & 27.07 & \textbf{0.846} & 0.178 \\ \bottomrule
\end{tabular}
}
\caption{Quantitative evaluation of the proposed sequence-to-sequence model with varying numbers of time points ($n=3,4,5$) in terms of normalized root mean squared error (NRSME), peak signal-to-noise ratio (PSNR), structural similarity \cite{Wang_Bovik_Sheikh_Simoncelli_2004}
    (SSIM) and the Spectral Angle Mapper (SAM) \cite{kruse1993spectral} metric. Our multi-temporal network with SAR guidance outperforms the multi-temporal ablation model without prior SAR information.}
    \label{tab:more_time}
\end{table}

\begin{table}[]
\scalebox{0.7}{
\begin{tabular}{@{}lllllllll@{}}
\toprule
\% cloud coverage          & NRMSE (all) & NRMSE (cloudy) & NRMSE (clear) & PSNR & SSIM & SAM  \\ \midrule
0-10 \% & \textbf{0.041} &  \textbf{0.046} & \textbf{0.041} & \textbf{28.59} & \textbf{0.870} & \textbf{0.143} \\
10-20 \%  & 0.044 & \textbf{0.046} & 0.043 & 27.69 & 0.848 & 0.166 \\
20-30 \% & 0.046 & 0.047 & 0.044 & 27.25 & 0.841 & 0.169 \\
30-40 \% & 0.048 & 0.050 & 0.045 & 26.77 & 0.830 & 0.169 \\
40-50 \% & 0.047 & 0.048 & 0.045 & 26.86 & 0.830 & 0.167 \\
50-60 \% & 0.049 & 0.494 & 0.048 & 26.55 & 0.825 & 0.185 \\
60-70 \% & 0.052 & 0.052 & 0.043 & 26.10 & 0.817 & 0.184 \\
70-80 \% & 0.049 & 0.050 & 0.044 & 26.59 & 0.816 & 0.179 \\
80-90 \% & 0.050 & 0.050 & 0.044 & 26.54 & 0.820 & 0.175 \\
90-100 \% & 0.063 & 0.063 & --- & 24.79 & 0.786 & 0.222 \\ \bottomrule
\end{tabular}
}
\caption{Performance of our sequence-to-point cloud removal method (n=3, with S1 \& with perceptual loss) as a function of cloud coverage. For a given interval, all n=3 input images are sampled to contain a corresponding extent of clouds. The outcomes show that image reconstruction performance is highly dependant on the percentage of cloud coverage. While performance decrease is not strictly monotonous with an increase in cloud coverage, a strong association persists.}
    \label{tab:cloudy_percentage}
\end{table}

\subsection{Sequence-to-Sequence Cloud Removal} \label{exp:seq2seq}

A key characteristic of training the sequence-to-sequence cloud removal model described in section \ref{subsec:seq2se2method} is the model being trained directly on the time series of images one aims to removes clouds from, without the use of any external training data as in \cite{ulyanov2018deep, zhang2019internal}. More specifically, the training procedure teaches the network to replicate cloud-free pixels and inpaint cloud-covered ones in the target sequence $S2$ according to the cost function $\mathcal{L}_{all}$ formulated in \cite{zhang2019internal} as

\begin{equation}
\label{eq:Lall_DIP}
\mathcal{L}_{all} = \lambda_{L2} \mathcal{L}_{L2} + \lambda_{perc} \mathcal{L}_{perc}
\end{equation}
\label{eq:Lall_seq2seq}

\begin{equation}
\label{eq:L2_DIP}
\mathcal{L}_{L2} = ||S2 \cdot (1-m), \hat{S2} \cdot (1-m)||_2
\end{equation}

\begin{equation}
\label{eq:Lperc_DIP}
\mathcal{L}_{perc} = ||VGG16(S2) \cdot (1-m), VGG16(\hat{S2}) \cdot (1-m)||_2
\end{equation}

where $\lambda_{L2} = 1$ and $\lambda_{perc} = 0.01$ refer to hyper-parameters that linearly combine the terms constituting $\mathcal{L}_{all}$. $\mathcal{L}_{2}$ is a pixel-wise reconstruction loss evaluated over the cloud-free pixels via an auxiliary VGG16 network \cite{Simonyan_Zisserman_2014} as explained before. The pseudo-code formalizing the intrinsic learning procedure is given in Algorithm \ref{algo:internal} described in section \ref{subsec:seq2se2method} and further justifications are stated in the original work of \cite{ulyanov2018deep}. For a given target sequence, the network is trained for 20 passes with batches of $n=5$ samples consisting of temporally adjacent images, for 100 iterations per pass. The network is optimized via ADAM \cite{kingma2014adam} with a learning rate of $0.01$  and the hyperparameters of Algorithm \ref{algo:internal} set as stated in \cite{zhang2019internal}.

To quantitatively evaluate the considered model on SEN12MS-CR-TS, we propose the following protocol for a sequence-to-sequence cloud removal task:
For a given target sequence, the least cloud-covered S2 observation is identified and denoted as a target image $S2_t$. The most cloudy S2 sample is observed and denoted as a source image $S2_s$. The cloud-covered pixels of $S2_s$ according to a cloud mask $m$ are alpha-blended with the cloud-free pixels of $S2_t$ similar to the approach of \cite{Rafique_Blanton_Jacobs}. Finally, the cloud-removed prediction $\hat{S2}_t$ is then compared against the originally cloud-free $S2_t$ in order to get a measure of goodness of cloud removal.

\begin{figure}[h!tb]
  \centering
  \begin{subfigure}[b]{0.99\linewidth}
    \includegraphics[width=\linewidth]{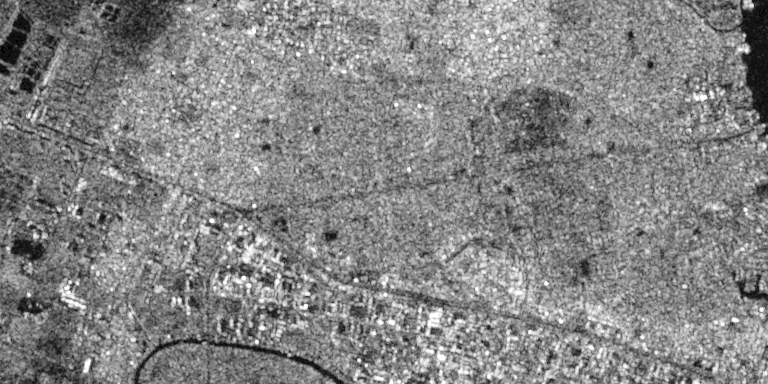}
  \end{subfigure}
    \begin{subfigure}[b]{0.99\linewidth}
    \includegraphics[width=\linewidth]{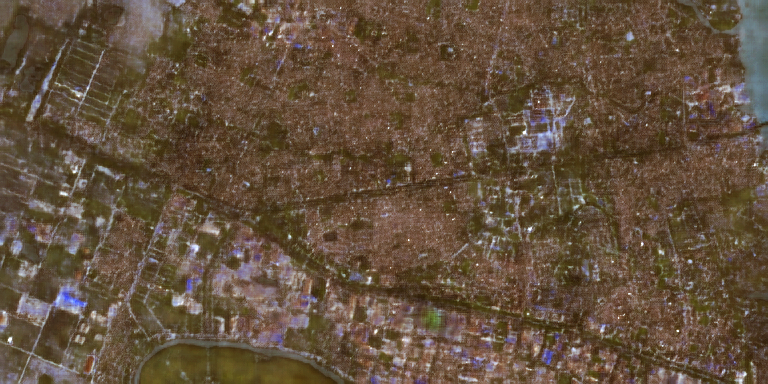}
  \end{subfigure}
  
  \begin{subfigure}[b]{0.99\linewidth}
    \includegraphics[width=\linewidth]{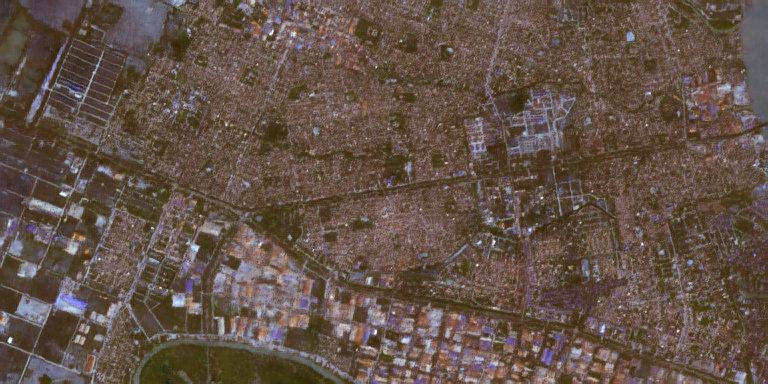}
  \end{subfigure}
    \begin{subfigure}[b]{0.99\linewidth}
    \includegraphics[width=\linewidth]{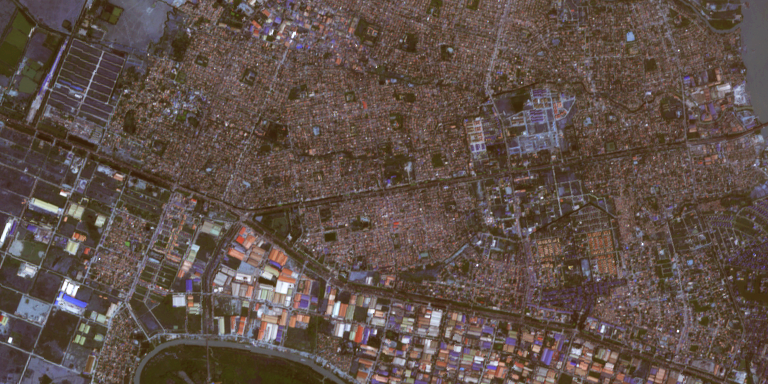}
  \end{subfigure}
  \caption{Illustrations on the effect of prior guidance via SAR information. Columns: SAR input to the SAR-conditioned model, cloud-free prediction of the model conditioned on Gaussian noise, cloud-free prediction of the model conditioned on SAR information, cloud-free observation as a reference image. The structural information provided by the SAR input provides a strong prior to the model, guiding it towards learning to remove clouds in the cloudy input time series.}
  \label{fig:ablation_seq2seq}
\end{figure}

Table \ref{tab:seq2seqmetrics} shows the results of the proposed network on the sequence-to-sequence cloud removal task following the aforementioned protocol. Furthermore, the considered model is compared against an ablation model, conditioned on random Gaussian noise as in \cite{ulyanov2018deep, zhang2019internal} in place of the meaningful $S1$ input observations. Example outcomes of sequence-to-sequence cloud removal on a given ROI are depicted in Fig. \ref{fig:teaser}. Furthermore, Fig. \ref{fig:ablation_seq2seq} provides a qualitative comparison between the predictions conditioned on SAR versus no prior information, underlining the benefits of multi-modal information. The results highlight that the internal learning approach can learn to reconstruct cloud-covered pixels on a very limited amount of data. Furthermore, the results demonstrate that including SAR data results in performance benefits over the single-sensor baseline.

\begin{figure*}[h!tb]
  \centering
    \label{fig:coffee}
  \begin{subfigure}[b]{0.32\linewidth}
    \includegraphics[width=\linewidth]{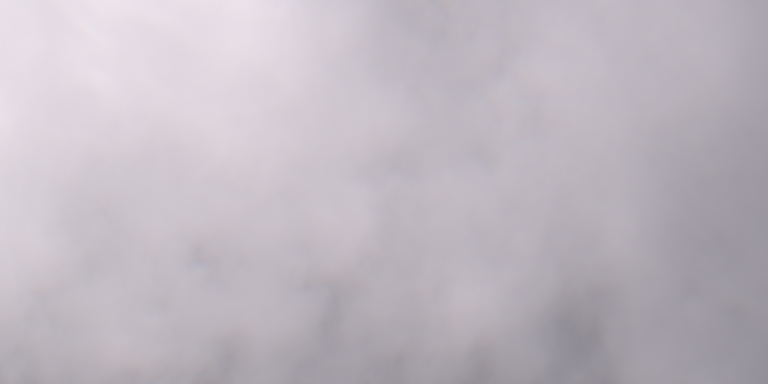}
  \end{subfigure}
  \begin{subfigure}[b]{0.32\linewidth}
    \includegraphics[width=\linewidth]{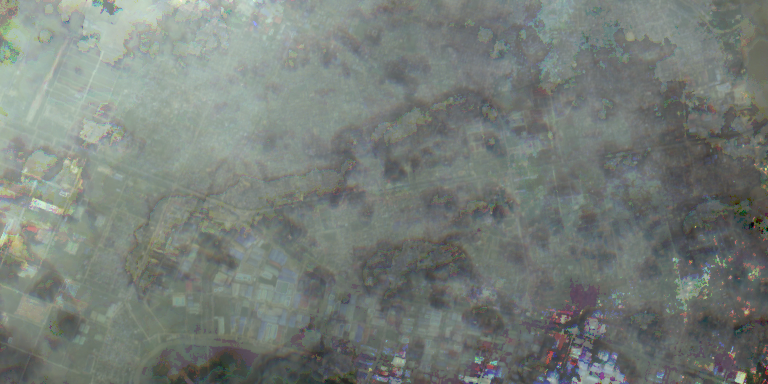}
  \end{subfigure}   
    \begin{subfigure}[b]{0.32\linewidth}
    \includegraphics[width=\linewidth]{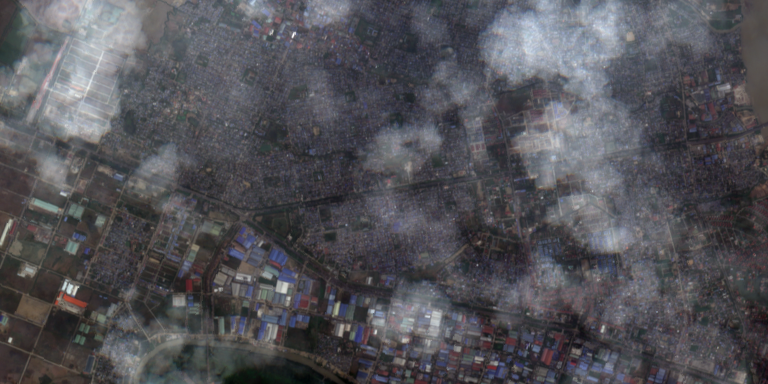}
  \end{subfigure}
      \begin{subfigure}[b]{0.32\linewidth}
    \includegraphics[width=\linewidth]{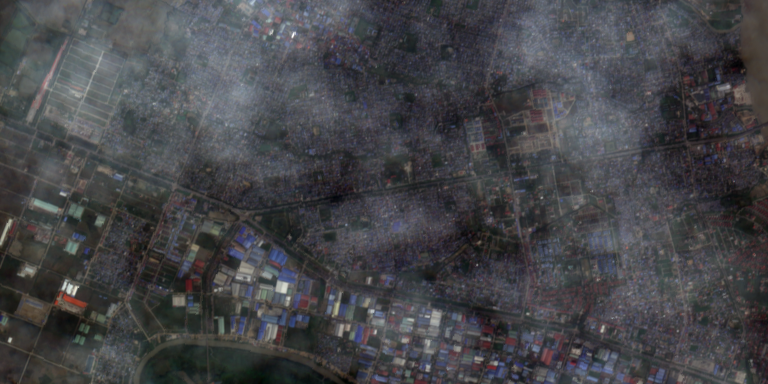}
  \end{subfigure}
  \begin{subfigure}[b]{0.32\linewidth}
    \includegraphics[width=\linewidth]{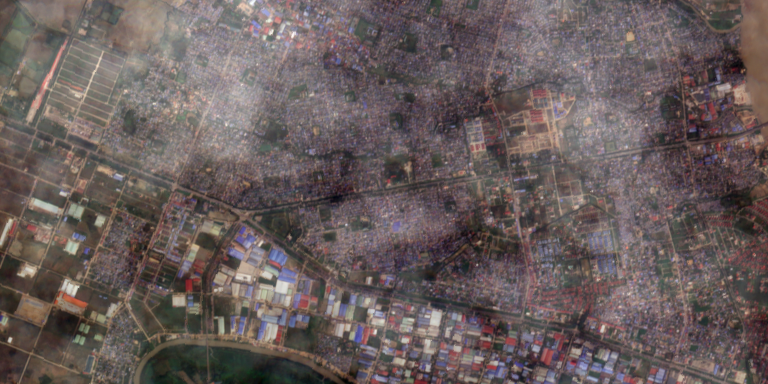}
  \end{subfigure}
    \begin{subfigure}[b]{0.32\linewidth}
    \includegraphics[width=\linewidth]{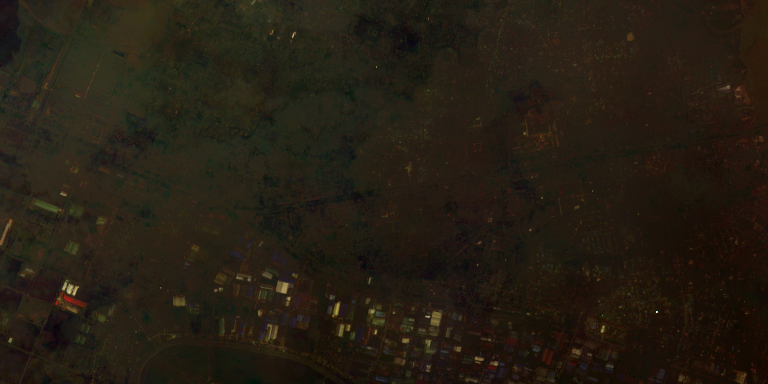}
  \end{subfigure}
  \caption{Illustration of baseline methods for the sequence-to-sequence cloud removal task. The presented results show a cloudy image to be de-clouded, as well as the predictions via Riemannian Robust Principal Component Pursuit (RPCP) \cite{hintermuller2015robust}, 
  Non-negative Matrix Factorization Incremental Subspace Learning (NMFISL) \cite{bucak2007incremental}, Probabilistic Non-negative Matrix Factorization (PNMF) \cite{lee2000algorithms}, Manhattan Non-negative Matrix Factorization (MNMF) \cite{guan2012mahnmf} and Online Stochastic Tensor Decomposition (OSTD) \cite{sobral2015online}. The results indicate that the presence of large and dense clouds poses a severe challenge for the considered methods. Most baselines de-cloud the image except for some residual artifacts, some techniques display discolorization. For comparison with ours (no S1), ours (with S1) and the cloud-free target image, see Fig. \ref{fig:ablation_seq2seq}}
  \label{fig:baselines_seq2seq}
\end{figure*}

\begin{table}[]
\centering
\begin{tabular}{@{}lllllll@{}}
\toprule
model             & NRMSE (all) & PSNR  & SSIM & SAM   \\ \midrule
RPCP \cite{hintermuller2015robust} & 0.403 & 7.911 & 0.264 & 30.567 \\
NMFISL \cite{bucak2007incremental} & 0.312 & 10.262 & 0.450 & 29.285 \\
PNMF \cite{lee2000algorithms} & 0.317 & 10.135 & 0.432 & 29.801 \\
MNMF \cite{guan2012mahnmf} & 0.361 & 8.945 & 0.361 & 28.685 \\
OSTD \cite{sobral2015online} & 0.303 & 10.853 & 0.402 & 35.454 \\
seq2seq (no S1)   & 0.298        & 11.434 & 0.494 & 28.127 \\
seq2seq (with S1) & \textbf{0.274}        & \textbf{11.590} & \textbf{0.512} & \textbf{27.733} \\ \bottomrule 
\end{tabular}
\caption{Quantitative evaluation of baseline methods and the proposed sequence-to-sequence model in terms of root mean squared error (RSME), peak signal-to-noise ratio (PSNR), structural similarity  
    (SSIM) and the Spectral Angle Mapper (SAM) \cite{kruse1993spectral} metric. Our multi-temporal network with SAR guidance outperforms the considered baselines as well as the multi-temporal ablation model without prior SAR information.}
    \label{tab:seq2seqmetrics}
\end{table}

\vspace{-1.5mm}

\section{Discussion} \label{discussion}

The main contribution of this work is in curating and providing SEN12MS-CR-TS, a multi-modal multi-temporal data set for cloud removal in optical satellite imagery. Our large-scale data set covers a heterogeneous set of ROI sampled from all over Earth, acquired in different seasons throughout the year. Given that the contained observations cover clear-view, filmy, as well as non-transparent dense clouds, the objective of reconstructing cloud-covered information poses a challenging task for the considered methods and future approaches. For the sake of demonstrating the usefulness of the presented data set, we propose a sequence-to-point as well sequence-to-sequence cloud removal network. The considered methods are evaluated in terms of pixel-wise and image-wise metrics. We provide evidence that taking time series information into account is facilitating the reconstruction of cloudy pixels and that including multi-sensor measurements does further improve the goodness of the cloud-removed predictions, justifying the design of SEN12MS-CR-TS to include multi-temporal and multi-modal data. The major difference to the preceding mono-temporal SEN12MS-CR data set \cite{ebel2020cloud} for cloud removal is that SEN12MS-CR-TS features a time series of 30 samples per ROI. This allows for developing methods that integrate information across time to more faithfully reconstruct cloud-obscured measurements. The sensitivity to temporal information may be particularly valuable for future research investigating the benefits of cloud removal to time-sensitive applications, such as change detection. On the other side, there is a trade-off in terms of size, and while SEN12MS-CR-TS is more than twice as large as its mono-temporal precursor, the latter contains about two times as many ROI sampled over all continents. However, both data sets are fully compatible; meaning that holdout ROI of one belong to the test split of the other data set and vice versa. As there is no geo-spatial overlap across splits between both data sets, they can be combined for training or validation purposes. Finally, the two data sets exhibit a comparable extent of cloud coverage---about 50 and 48 \% respectively, both covering the full spectrum from semitransparent haze to thick and dense clouds. A discrepancy between both data sets is in SEN12MS-CR having between 25 and 50 \% overlap between neighboring patches (following the design of \cite{schmitt2019sen12ms}), whereas SEN12MS-CR-TS has no intersection between adjacent samples. SEN12MS-CR contains 122,218 patch triplets of S1, cloudy S2 and cloud-free S2 data, whereas SEN12MS-CR-TS consists of 30 time samples for each of the 15578 patch-wise observations, for every S1 and S2 measurement. Due to the differences in pre-processing the two data sets are not co-registered patch-wise but, importantly, they share a common definition of ROI as well as train and test splits. This way, they are compatible with one another such that SEN12MS-CR-TS can be utilized for time-series cloud removal while SEN12MS-CR can provide further geospatial coverage of additional ROI on individual time points. Thanks to the different designs of both data sets, they may prove beneficial facilitating a variety of downstream tasks, such as semantic segmentation \cite{schmitt2019sen12ms}, scene classification \cite{schmitt2021remote} or change detection \cite{ebel2021fusing}, even in the presence of clouds.

Beyond the design of our novel data set, additional contributions of this work are in introducing the internal learning approach to cloud removal in optical satellite data, as well as demonstrating that SAR-to-optical cloud removal performs better than the original noise-to-optical translation framework. While our data set aims to provide a global distribution of samples, we think that the internal learning approach to cloud removal may be of particular interest for remote sensing practitioners focusing on a single a spatially confined ROI, as no further external data is necessary. 

\section{Conclusion} \label{conclusion}

As a large extent of our planet is covered by haze or clouds at any given point in time, such atmospheric distortions pose a severe constraint to the ongoing monitoring of Earth. To approach this challenge, our work presented SEN12MS-CR-TS, a multi-modal and multi-temporal data set for training and evaluating global and all-season cloud removal methods. Our data set contains Sentinel-1 and Sentinel-2 observations from over $80,000$ $km^2$ of landcover, distributed globally and recorded through the year. The globally distributed ROI each are large-sized, and capture a heterogeneous mass of landcover. We demonstrated the practicality of SEN12MS-CR by considering two methods: First, a model for sequence-to-point cloud removal. Second, a network for sequence-to-sequence cloud removal which, to our knowledge, provides the first case a model preserving temporal information is proposed in the context of cloud removal. Both methods benefited from the presence of co-registered and paired SAR measurements contained in our data set. The conducted experiments highlight the contribution of our curated data set to the remote sensing community as well as the benefits of multi-modal and multi-temporal information to reconstruct noisy information. SEN12MS-CR is made public to facilitate future research in multi-modal and multi-temporal image reconstruction.

\section*{Acknowledgment}

The authors would like to thank ESA and the Copernicus program for making the Sentinel observations accessed for this submission publicly available. Furthermore, we would like to thank Rewanth Ravindran for assisting us in the data curation process.

\ifCLASSOPTIONcaptionsoff
  \newpage
\fi

\appendices

\section{Temporal coincidence of paired observations} \label{appendix:a}

\begin{figure}[h!tb] 
    \includegraphics[width=\linewidth]{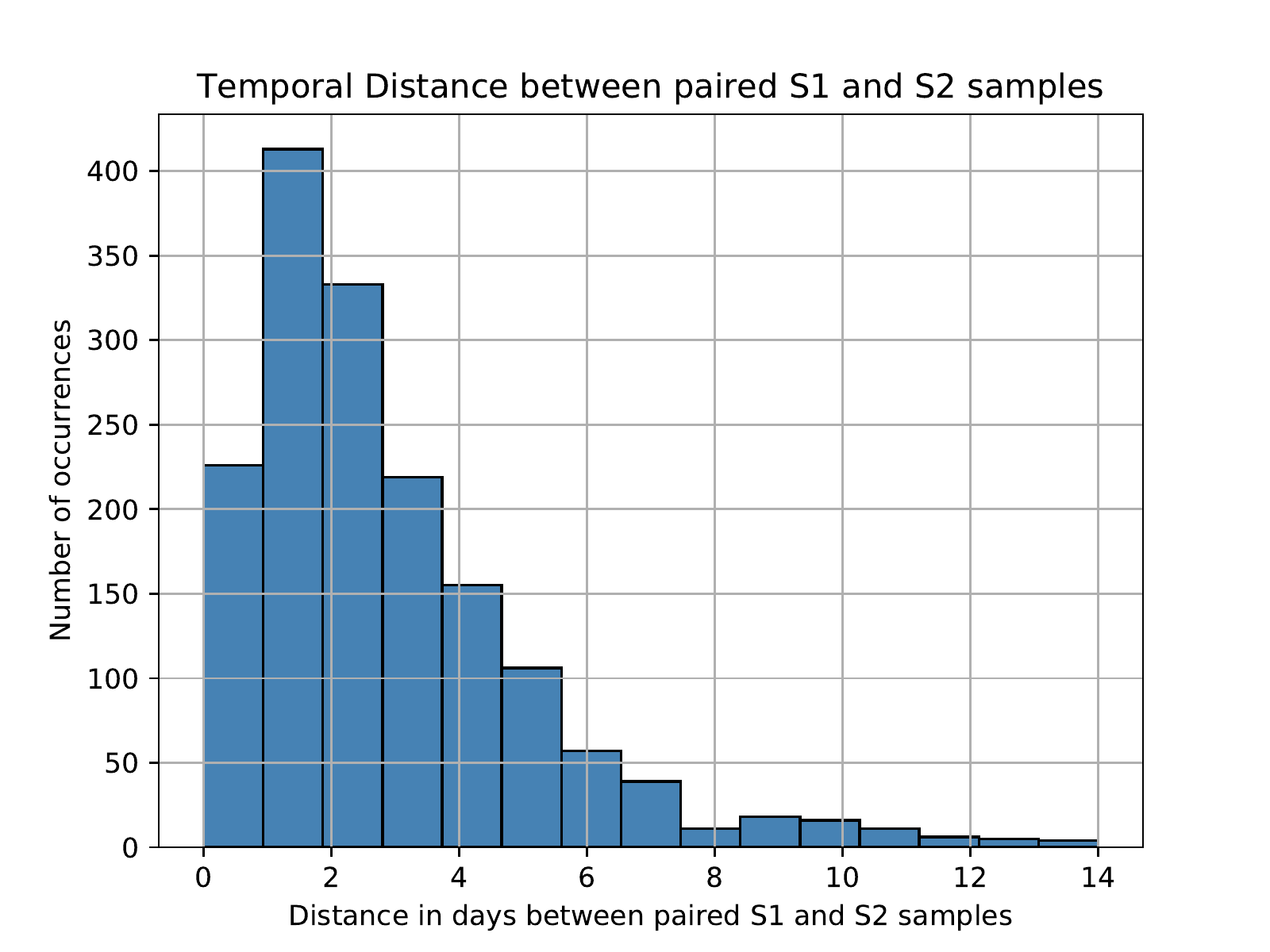}
    \caption{Histogram of temporal differences between paired observations. The mean time differences across all paired observations is $2.61$ ($\pm$ $2.41$), indicating a close proximity between paired samples.}
    \label{fig:temp_coincidence}
\end{figure}

Full-scene observations of Sentinel-1 and Sentinel-2 are collected within a 14-days time window in a paired manner, as specified in section \ref{curation}. To further analyze the temporal distance within paired data, Fig. \ref{fig:temp_coincidence} illustrates the empirically observed coincidences within SEN12MS-CR-TS. The mean time differences across all paired observations is $2.61$ ($\pm$ $2.41$), which is considerably smaller than the interval bound and implies a close proximity between paired samples.

\bibliographystyle{IEEEtran}
\bibliography{references}

%



%

\begin{IEEEbiography}[{\includegraphics[width=1in,height=1.25in,clip,keepaspectratio]{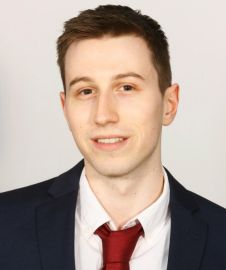}}]{Patrick Ebel} (patrick.ebel@tum.de) received the B.Sc. degree in cognitive science from University of Osnabrück, Germany, in 2015. He received the M.Sc. degree in Cognitive Neuroscience and the M.Sc degree in Artificial Intelligence from Radboud University Nijmegen, The Netherlands, in 2018. He is working as doctoral candidate in the Data Science in Earth Observation lab, Department of Aerospace and Geodesy of Technical University of Munich (TUM), Germany. His research interests include machine learning as well as its applications in computer vision and remote sensing. Specifically, he is working on multi-modal and multi-temporal data fusion and automated image reconstruction methods. \end{IEEEbiography}

\begin{IEEEbiography}[{\includegraphics[width=1in,height=1.25in,clip,keepaspectratio]{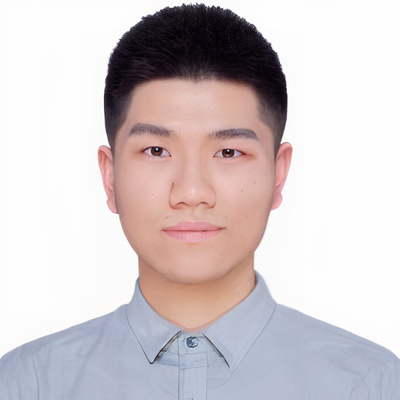}}]{Yajin Xu} (yajin.xu@tum.de) received the B.Sc. degree in engineering with distinctions from Wuhan University, Wuhan, China in 2018 and is pursuing an M.Sc. double degree in Earth Oriented Space Science and Technology (ESPACE) at Wuhan University, Wuhan, China and Technical University of Munich (TUM), Munich, Germany. His study focus is machine learning applied to remote sensing data. In 2021 he was a research assistant in the Data Science in Earth Observation (SiPEO) group at TUM and Remote Sensing Technology Institute of DLR, investigating deep learning-based approaches for cloud removal in optical satellite data. His interests are in geospatial data analysis. \end{IEEEbiography}

\begin{IEEEbiography}[{\includegraphics[width=1in,height=1.25in,clip,keepaspectratio]{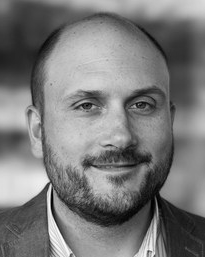}}]{Michael Schmitt}(michael.schmitt@unibw.de) received his Dipl.-Ing. (Univ.) degree in geodesy and geoinformation, his Dr.-Ing. degree in remote sensing, and his habilitation in data fusion from the Technical University of Munich (TUM), Germany, in 2009, 2014, and 2018, respectively.
Since 2021, he has held the Chair for Earth Observation at the Department of Aerospace Engineering of the Bundeswehr University Munich in Neubiberg, Germany. Before that, he was a professor for applied geodesy and remote sensing at the Munich University of Applied Sciences, Department of Geoinformatics. From 2015 to 2020, he was a senior researcher and deputy head at the Professorship for Data Science in Earth Observation at TUM; in 2019 he was additionally appointed as Adjunct Teaching Professor at the Department of Aerospace and Geodesy of TUM. In 2016, he was a guest scientist at the University of Massachusetts, Amherst. His research focuses on image analysis and machine learning applied to the extraction of information from multi-modal remote sensing observations. In particular, he is interested in remote sensing data fusion with a focus on SAR and optical data. He is a co-chair of the Working Group ``SAR and Microwave Sensing'' of the International Society for Photogrammetry and Remote Sensing, and also of the Working Group ``Benchmarking'' of the IEEE-GRSS Image Analysis and Data Fusion Technical Committee. He frequently serves as a reviewer for a number of renowned international journals and conferences and has received several Best Reviewer awards. He is a Senior Member of the IEEE and an associate editor of IEEE Geoscience and Remote Sensing Letters.
\end{IEEEbiography}

\begin{IEEEbiography}[{\includegraphics[width=1in,height=1.25in,clip,keepaspectratio]{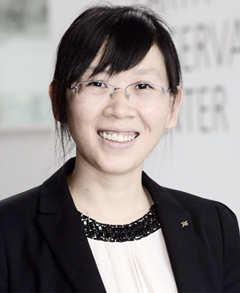}}]{Xiao Xiang Zhu}
(S'10--M'12--SM'14--F'21) received the Master (M.Sc.) degree, her doctor of engineering (Dr.-Ing.) degree and her “Habilitation” in the field of signal processing from Technical University of Munich (TUM), Munich, Germany, in 2008, 2011 and 2013, respectively.
\par
She is currently the Professor for Data Science in Earth Observation (former: Signal Processing in Earth Observation) at Technical University of Munich (TUM) and the Head of the Department ``EO Data Science'' at the Remote Sensing Technology Institute, German Aerospace Center (DLR). Since 2019, Zhu is a co-coordinator of the Munich Data Science Research School (www.mu-ds.de). Since 2019 She also heads the Helmholtz Artificial Intelligence -- Research Field ``Aeronautics, Space and Transport". Since May 2020, she is the director of the international future AI lab "AI4EO -- Artificial Intelligence for Earth Observation: Reasoning, Uncertainties, Ethics and Beyond", Munich, Germany. Since October 2020, she also serves as a co-director of the Munich Data Science Institute (MDSI), TUM. Prof. Zhu was a guest scientist or visiting professor at the Italian National Research Council (CNR-IREA), Naples, Italy, Fudan University, Shanghai, China, the University  of Tokyo, Tokyo, Japan and University of California, Los Angeles, United States in 2009, 2014, 2015 and 2016, respectively. She is currently a visiting AI professor at ESA's Phi-lab. Her main research interests are remote sensing and Earth observation, signal processing, machine learning and data science, with a special application focus on global urban mapping.

Dr. Zhu is a member of young academy (Junge Akademie/Junges Kolleg) at the Berlin-Brandenburg Academy of Sciences and Humanities and the German National  Academy of Sciences Leopoldina and the Bavarian Academy of Sciences and Humanities. She serves in the scientific advisory board in several research organizations, among others the German Research Center for Geosciences (GFZ) and Potsdam Institute for Climate Impact Research (PIK). She is an associate Editor of IEEE Transactions on Geoscience and Remote Sensing and serves as the area editor responsible for special issues of IEEE Signal Processing Magazine. She is a Fellow of IEEE.
 \end{IEEEbiography}







\end{document}